\documentclass[preprint,12pt]{elsarticle}

\usepackage{graphicx}
\usepackage{multirow}
\usepackage{subcaption}
\usepackage{natbib}

\usepackage{natbib}
 \bibpunct[, ]{(}{)}{,}{a}{}{,}%
 \def\bibfont{\small}%
 \def\bibsep{\smallskipamount}%
\journal{Journal of Econometrics}

\begin{document}

\begin{frontmatter}

\newcommand{\todo}[1]{\textbf{TODO: #1}}

\title{Sequence embeddings help to identify fraudulent cases in healthcare insurance}

\author[sk]{I. Fursov}
\author[sk]{A. Zaytsev}
\author[sk]{R. Khasyanov}
\author[hamb]{M. Spindler}
\author[sk]{E. Burnaev}

\address[sk]{Skoltech}
\address[hamb]{University of Hamburg}

\begin{abstract}
Fraud causes substantial costs and losses for companies and clients in the finance and insurance industries. Examples are fraudulent credit card transactions or fraudulent claims. It has been estimated that roughly $10$ percent of the insurance industry's incurred losses and loss adjustment expenses each year stem from fraudulent claims. The rise and proliferation  of digitization in finance and insurance has lead to big data sets, consisting in particular of text data, which can be used for fraud detection. In this paper we propose architectures for text embeddings via deep learning, which help to improve the detection of fraudulent claims compared to other machine learning methods. We illustrate our methods using a data set from a large international health insurance company. The empirical results show that our approach outperforms other state-of-the-art methods and can help make the claims management process more efficient. As (unstructured) text data become increaslingly available to economists and econometricians, our proposed methods will be valuable for many similar applications, particularly when variables have a large number of categories as is typical for example of the International Classification of Disease (ICD) codes in health economics and health services.
\end{abstract}


\begin{keyword}
embeddings \sep 
deep learning \sep 
fraud detection \sep 
structured data \sep 
health insurance \sep 
social media and text
\end{keyword}

\end{frontmatter}

\section{Introduction}
\label{sec1}

Fraud causes substantial costs and losses for companies and clients in the finance and insurance industries. Examples include fraudulent credit card transactions or insurance claims. 
Indeed, it has been  estimated that roughly $10$ percent of the insurance industry's incurred losses and loss adjustment expenses each year stem from fraudulent claims.\footnote{\texttt{https://www.iii.org/article/background-on-insurance-fraud}}  
Hence fraud detection is a key function in these industries and core to the claims management process. Detecting fraud is also considered as a key competence of insurance and finance companies.

The rise and proliferation of digitization in finance and insurance have led to big data sets, which can be exploited for fraud detection. 
In this paper, we propose architectures for text embeddings via deep learning, that help improve the detection of fraudulent claims compared to other machine learning methods. 
We illustrate our method using a data set from a large international health insurance company. 


Analyzing fraud with statistical and machine learning methods poses special challenges. First, transaction data and claims data are often available only in a so-called unstructured format. Second, fraud data are highly unbalanced, meaning that the number of fraudulent cases is very small compared to the number of non-fraudulent ones. This fact influences the choice of the classification approach and performance measures. Third, claims do not have a fixed length because the number of items in an invoice varies. One approach to this might be to equalize the input length by filling with zeros, but doing so often leads to distorted results. 
It is well known that deep learning outperforms other machine learning  methods for analyzing unstructured data comprising, for example, text or images. In this paper, we develop deep learning architectures that are tailored to claims data and can handle each of the challenges listed above when processing unstructured information. 

Our analysis is based on doctor's bills, which have an interesting structure that is common to many economic data sets, in particular those used to address microeconomic problems. Such bills usually consist of unstructured text and have the properties of text data, for example insofar as the number of claims/items varies from bill to bill. Unlike text data, however, the exact ordering of the claims in each bill is irrelevant. It is also typical of doctor's bills that some variables are coded with many thousands of categories. In this paper, we develop methods for analyzing such (semi)unstructured data, that might also be useful for many other applications.

We test our methods on a data set from a health insurance company. Our  empirical results show that these outperforms other state-of-the-art methods in predicting fraudulent claims, and help make the claims management process more efficient.

\textbf{Plan of the paper:} After the introduction (Section \ref{sec1}), we provide an overview of the literature and state-of-the-art methods (Section \ref{sec2}). In Section \ref{sec3}, we present models and methods for analyzing general text data and for analyzing the special structure of claims data to detect insurance fraud. In the next Section \ref{sec4}, we describe our data set, and in Section \ref{sec5} we present the results of our analysis. Finally, we present our conclusions in Section \ref{sec6}.
   
\section{Overview of the literature}
\label{sec2}

\subsection{Anomaly detection}

Detecting anomalies in data is one of the core problems in data analysis and has been investigated in recent years  within diverse research areas and application domains, including time-series modeling (see \cite{QuasiPeriodic,kNN2017}), predictive maintenance of technical systems (see \cite{Degradation2016,OCSVM2018}), and applications in the finance and insurance industries (see \cite{Chandola}). 

Anomalies in data are important because they can translate to significant and often critical, actionable information in a wide variety of application domains. For example, in credit card transactions, anomalies can indicate when unauthorized purchases have taken place as a result of credit card or identity theft, see \cite{Jurgovsky}. Similarly, anomalies in health insurance claims can be indicative, for example, of  deception of  misrepresentation carried out to gain an inappropriate or unjustified health benefit, or of billing for services not rendered, see \cite{kirlidog}.

We also refer to the reviews of~\cite{zhou2018state,phua2010comprehensive}, dedicated to various fraud detection problems and machine-learning based solutions to them. 

\subsection{Machine learning for healthcare and insurance}

While a number of machine learning methods have been applied to problems in healthcare and insurance in recent years, deep learning and embeddings for fraud detection do not seem to have been covered by the literature until very recently. One of these few examples can be found in a recent study by \cite{wang2018leveraging}, who focused on the detection of automobile insurance fraud.

\cite{wang2018leveraging} focused in their recent study, on the detection of automobile insurance fraud. They processed text descriptions of the accidents, extracting traditional text features manually and combined these with features, extracted automatically using deep learning. While their model showed accuracy superior to that achieved using existing approaches, neither the precise architecture of the best model nor the approach to training and validating the model is described clearly in their paper.

Another example can be seen in a recent article by~\cite{kim2019fraud} who used hierarchical clustering based on deep neural networks to detect fraud in descriptions of candidates during job recruitment, significantly improving the accuracy of detection compared to conventional methods.

In order to predict instances of automobile insurance frauds~\cite{balasubramanian2019ensemble} used manually-crafted features.

\subsection{Construction of embeddings}

Embeddings for anomaly detection problems have been used in different application domains to solve anomaly detection and various other problems. For example, \cite{chen2016entity} developed an approach to embed entities, representing events in real computer systems, into a common latent space. Each event involved heterogeneous types of attributes: time, user, source process, destination process, and so on. \cite{Hu2016AnEA} studied the problem of detecting structurally inconsistent nodes in graphs, for example to detect outlier authors in a network in which  different authors were connected if co-authored a paper. 

However, it is embeddings constructed for applications related to natural language processing that are currently attracting the most attention. We focus here on papers with embeddings of simple entities, such as words. 
These include the classic TF-IDF approach, explained for instance  by~\cite{rajaraman2011mining}, and the recent and well-known word2vec by~\cite{mikolov2013distributed} and GloVE by~\cite{pennington2014glove}. The latter two methods take into account the concurrences of words, while TF-IDF is simply a normalized one-hot-encoding for a dictionary of words at hand. 

To unite embeddings of simple entities, there are also a number of approaches. 
For example, we are able to construct an embedding of a text from an embedding of each word within this text.
Simple heuristics include taking maximum value among each dimension for embeddings of words or taking mean values, see~\cite{arora2016simple}.
More complex approaches are based on convolutional and recurrent neural networks, see~\cite{wang2016cse,kiros2015skip,arora2016simple}.

\subsection{Imbalanced classification problems}
\label{sec:imb_classificaion}

Skewed distribution (imbalanced classes) is considered one of the most critical challenges to solving fraud detection problems. Generally speaking, there are far fewer instances of fraudulent items than normal ones. The resulting imbalance makes it difficult for learners to detect  patterns in the minority class data. There are currently three main approaches to learn from imbalanced data \cite{Krawczyk2016}:
\begin{itemize}
    \item Data-level methods that modify the data set to balance their distributions and/or remove difficult observations
    \item Algorithm-level methods that directly modify existing learning algorithms to alleviate the bias towards majority objects and adapt them to mining data with skewed distributions,
    \item Hybrid methods that combine the advantages of these two approaches
\end{itemize}

In the data-level approach, \cite{duman} used under-sampling for a skewed class in a credit card fraud detection system, and \cite{Imbalance2015} assessed how resampling multiplier selection influences on classification accuracy. In the  algorithmic-level approach, \cite{Sahin2013} used cost-sensitive classifiers to address the class imbalance problem. In turn, \cite{fraudminer} proposed  the FraudMiner model, which is able to handle class imbalance by entering the unbalanced data directly to classifier.
More general-purpose approaches include over-sampling, see~\cite{chawla2002smote}, combinations of over- and under-sampling, see~\cite{saez2015smote}, and meta-learning to  automate selection of imbalanced classification methods, see \cite{Imbalance2019}.

\section{Methods}
\label{sec3}

\subsection{Learning of classical data-based models}

The common scenario for supervised learning is the following: we have a sample of observations, each containing a description of an object (given by features) and values of target variables for that object.
In disability insurance, for example, annual income, education, occupation, age and past medical records make up a description of a customer, and the target variable is a binary label signifying whether the customer's claim is fraudulent (fraud) or justified (not fraud).

In case of our data set from a large international health insurance company, the observations consist of doctor's bills that include  information about the treatments provided, their costs and dates, and final amounts. The target variable is whether the bill was classified as a fraud by a clerk handling the claim.

\begin{figure}
    \centering
    \includegraphics[width=0.7\textwidth]{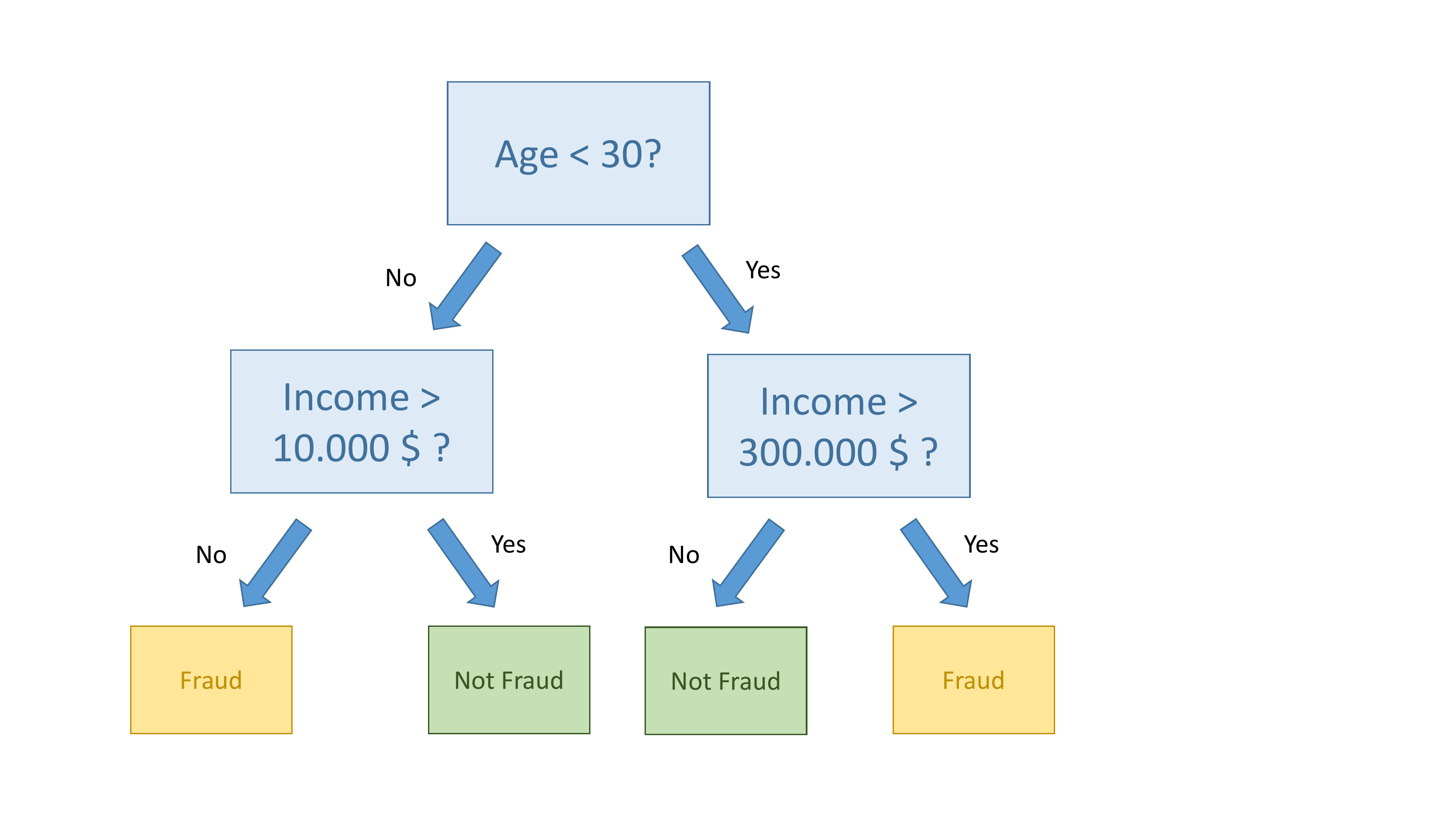}
    \caption{A scheme of a decision tree for fraud detection in the disability insurance: we start at the root node and go down making decisions on the directions, selected according to features of the object. At leaf nodes we make a decision according to their labels. The scheme does not represent a real model.}
    \label{fig:decision_tree_example}
\end{figure}

Thus, we can learn a model that predicts the target variables by taking features of a new object as its input. An example of a widely adopted model is a decision tree. In Figure~\ref{fig:decision_tree_example}, we provide an example of a decision tree for some input features. 

The power of machine learning is that we can learn a model that adapts to the given sample, and also generalizes well to unseen data that are similar to the training sample. Machine learning methods makes it possible to learn non-linear and complex relationships in data sets.

The common limitation of classic approaches is that they require the descriptions of objects to be in a restrictive format. Usually they use vectors of fixed, small length, which is not the case for many real-world objects such as texts or images with millions of pixels. In the case of insurance, the texts have different lengths and describe visits to a doctor with each patient having a different number of visits or doctor's bills and the invoice listing a varying number of treatments.


Data scientists devised various ways to generate features from complex, but structured data such as images and texts, in a manual fashion. These manually generated features are used as input for classic machine learning models. In economic applications, manual feature generation is also widely used -- for example the variable ``age''  is often constructed out of the variable ``date of birth''. Such approaches yield results of reasonable but limited quality.

\subsection{Deep learning revolution}

\begin{figure}
    \centering
    \includegraphics[width=\textwidth]{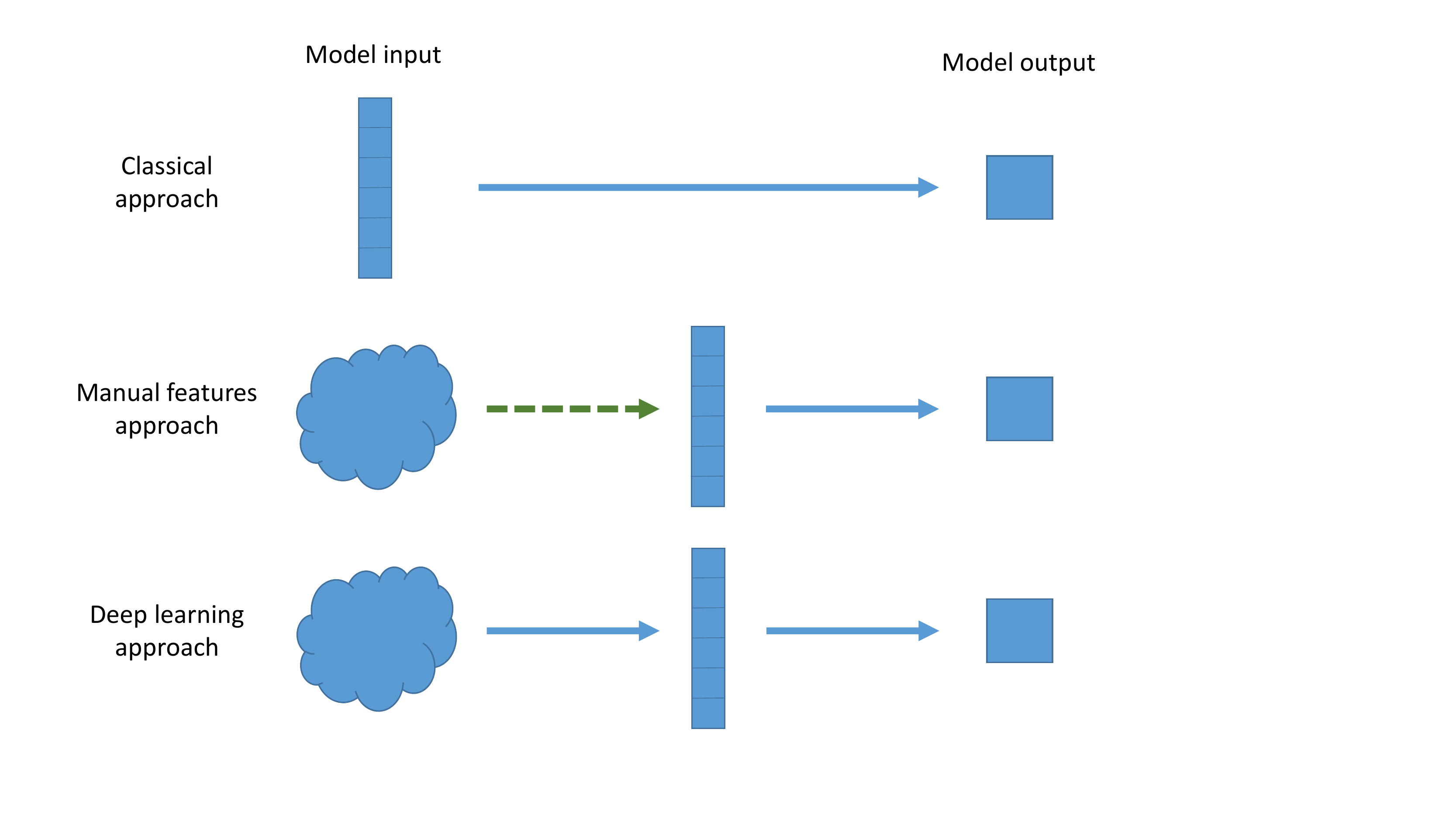}
    \caption{Classical approaches cannot handle complex weakly structured descriptions of objects. In the past, features from structured descriptions of objects were produced manually. Now, deep learning approaches produce such features in an automatic way.}
    \label{fig:comparison_approaches}
\end{figure}

The deep learning revolution changed the rules of the game in machine learning, data based models ~\cite{makridakis2017forthcoming}. 
Now algorithms can learn representations or embeddings of object descriptions to generate features that are informative enough to provide  accurate predictions while using rather simple machine learning models, such as fully-connected neural networks with only a few layers or decision trees. In summary, the strength of deep learning lies in feature extraction, which means learning informative features from high-dimensional, unstructured and complex input data. A schematic comparison of the classic approach, the classic approach with manually generated features and the deep learning approach is presented in  Figure~\ref{fig:comparison_approaches}.

The three driving forces behind the deep learning revolution are the  availability of new algorithms (e.g., the convolutional neural networks of~\cite{lecun2015deep} for image processing, recurrent neural nets for sequences, and word embeddings for texts), new hardware (graphical processing units, see~\cite{vasilache2014fast}), and vast samples of data (e.g. ImageNet dataset contains more than $14$ million of labeled images, see~\cite{ILSVRC15}).

The most successful application of deep learning is in the field of image processing.
However, sound advancements in deep learning have also been achived in  recognition, see~\cite{amodei2016deep}; for natural language processing (NLP), see~\cite{young2018recent}, and for graph data, see~\cite{hamilton2017representation}.  

The key idea of deep learning is to apply a sequence of nonlinear transformations (layers of the neural network)
of the object description to produce an informative embedding and use it as input to a final classifier.

\subsection{Concept of embeddings}

In this paper, we address the problem of representing healthcare insurance data using embeddings for the purpose of fraud detection. Embedding is a transformation of object descriptions to vectors that belong to the same low-dimensional space. Instances of these low-dimensional representations are such that the instances that are more alike have a smaller distance between them in the embedded space. For example, a good embedding provides vector representations of words in such a way that the relationship between two vectors mirrors the relationship between the two words. A popular word2vec model that has proven its effectiveness  in natural language processing tasks, constructs a low-dimensional vector of real numbers such that words appearing in a similar context have similar vector representations. 

In our case, we learn an embedding space constructed specifically for sequential data from healthcare insurance claims. Such representation significantly helps to detect fraudulent patterns. Thus, embedding is first  a general framework for dimension reduction and, second, an effective approach to extracting the features of intrinsic relations between objects.

To make these ideas clear, suppose that a text consists of words that come from a pool of $V$ different words (a so-called dictionary). One way to transform the text into numeric features would be to one-hot encode each word of the text so that each word in the text sequence is represented as a $V$-dimensional vector consisting of zeros except one entry at the location corresponding to that word. This is a standard way of encoding categorical variables. The representation is not efficient, however, if the dictionary is large, which is a typical situation not only for general texts, but also for healthcare data.
In turn, embeddings of words from the dictionary are represented  by real-valued $s$-dimensional vectors, such that $s$ is much lower than the size of $V$. This allows a compressed representation of the input textual description. In this representation usually the entries of the embedded vector are usually all different from zero. The embedding of the dictionary into the vector space should also maintain some relations between words. For example, a desirable property of the word embeddings is that the difference in the vector space between words ``queen'' and ``king'' should be similar to that between the words ``woman'' and ``man''. Learning word embeddings with such properties makes them a very powerful tool for text analysis.

\subsection{Application of embeddings}

In recent years, learning embeddings to represent complex relationships in data has become a common approach in the machine learning community. As a result, different types of embeddings have been used in many domains, such as natural language processing (NLP), network analysis, and computer vision. 

Word embeddings, such as word2vec by~\cite{mikolov2013efficient}, GloVe  by~\cite{pennington2014glove}, AdaGram  by~\cite{bartunov2016breaking} and others, provide vector representations of words such that the relationship between two vectors mirrors some linguistic relationship between the two words. 	
In supervised problems, word and sentence embeddings have proven effective in natural language processing tasks such as part-of-speech tagging, see \cite{collobert2011natural}, phrase-based machine translation, see \cite{zou2013bilingual}, named-entity recognition, see \cite{ma2016end}, and word sense disambiguation, see \cite{bartunov2016breaking}.

Graph and network embeddings attempt to capture local and global attributes on graphs, either based on engineered graph features or driven by training on graph data. Classical approaches to graph embeddings include feature-based methods such as graph kernels, see \cite{rwkernel:10, rconvolution}, and data-driven algorithms that yield distributed graph representations, see \cite{deepgraph,learncnn:16,AWE}. Using such embeddings we can solve various tasks related to network data analysis -- for example \cite{InfluenceSet2018} used anonymous walk  embeddings for graph influence set completion.

\section{Model}
\label{sec4}

We process data in a way similar to that shown in  Figure~\ref{fig:comparison_approaches}. Thus, we need to specify which method we use to generate features from initial descriptions of objects, and how we train machine learning models that predict whether the treatment is fraudulent or not.

In the following subsections we provide details about each step. When using neural networks, feature generation and model training occur simultaneously, so we describe both of these steps in the related subsection.

\subsection{Generation of features and embeddings}

For gradient boosting, we used BoW (bag of words) and TF-IDF (term frequency-inverse document frequency)  representations of the sequence of treatments. The idea behind BoW is to represent a text by counting the number of times $n_{w, t}$ a word $w$ is used in a text $t$. This gives a vector of frequencies of the words in the dictionary. Because words like ``a'', ``the'' show up many times, but provide less information, a normalized versions of BoW leads to TF-IDF.
We get a TF-IDF representation of the text as the product of the term frequency and inverse document frequency. The frequency term is equal to $n_{w, t}$ divided by the total number of words in the text $\sum_{w'} n_{w', t}$. The inverse document frequency is equal to the logarithm of the total number of documents divided by the number of documents that contain the considered word $w$. 
Note, that neither of these approaches take account of the order of words in a text; neglecting the order can decrease performance in many problems.

For approaches based on neural networks we use the same  vocabulary. Hence, the size of the input to the machine learning approach is equal to $|V| + g$, where $g$ is the number of additional features and $|V|$ is the vocabulary size.

Embedding matrix $E$ comes from the \textbf{word2vec} model, available in \textbf{Gensim} package~\cite{rehurek_lrec}, with $window\_size =10$, $min\_count =2$ and other hyperparameters set to default values. The embedding matrix contains in its columns for each word  of the dictionary the corresponding representation in the embedded vector space.

\subsection{Model training}

\subsubsection{Gradient boosting}

In machine learning the most common model for classification aside from neural networks is ensembles of decision trees, see \cite{fernandez2014we}. 
For each separate decision tree, we pass through it for a given object according to the values of input variables at each node until reach a leaf;
in a leaf the classifier returns the probabilities to belong to classes.
In an ensemble, we use a weighted sum of basic decision tree classifiers.

We adopt a gradient boosting algorithm to construct ensembles of decision trees (see \cite{chen2016xgboost}) as an easy-to-use approach that provides state-of-the-art performance in many problems. 
The algorithm has the following main hyperparameters: 
the number of trees in the ensemble, the maximum depth of each tree, 
the share of features used in each tree, the share of samples used for training of each tree, and the learning rate. 

Ensembles of decision trees are fast to construct, almost avoid over-fitting, successfully handle missing values and outliers and provide competitive performance, see \cite{fernandez2014we}.
One of the many advantages of gradient boosting is its ability to solve imbalanced classification problems and easy incorporate various imbalanced classification heuristics, see \cite{kozlovskaya2017deepboost}.

\subsubsection{Deep learning approach}
\label{aggr}

Our main deep learning approach is simple word-embedding-based models (SWEMs) from ~\cite{swem} that show strong performance in many natural language processing tasks. 
Below, we describe each layer of these models in more details. 

\begin{enumerate}
    \item \textbf{Treatment Embedding Layer} maps each treatment (single item of the doctor's bill) to a vector space using a trainable embedding matrix $E$ of size $|V| \times d$, where $V$ is a vocabulary set. Here it is the set of all billable treatments.  We pre-trained this layer using \textbf{Word2Vec} and used its weights to initialize  the embedding matrix. Let $\{t_1, t_2,\cdots, t_k\}$ represent the treatments in the input sequence (doctor's bill) of size $k$. Each treatment in a sequence is embedded into a vector of side $d$. Therefore, for each record (bill) we obtain a treatments-feature matrix $X \in \mathbf{R}^{k \times d}$. Another option to perform embeddings is through various neural recurrent architectures like LSTM, GRU, etc
    \item \textbf{Aggregation Layer} aggregates the sequence of embeddings using element-wise average, taking maximum values or concatenating along each dimension over the treatments vectors. This layer combines information about each treatment into a single vector.



  \item \textbf{Several Multilayer Perceptron} (MLP) layers with ReLU activations. Each layer learns features from the sequence of treatments at different levels of granularity. We expect the model to pay attention to the features that indicate the possibility of fraud.
  \item \textbf{Extra tower} is used to take additional features into account. We included extra tower with fully-connected layers over meta-features (gender, age, insurance type, etc.). The outputs of the treatment tower and the feature tower are concatenated before passing through two fully connected layers with ReLU activations.
  \item \textbf{Output layer} is represented by an MLP layer is followed by a softmax function to receive class probabilities.
\end{enumerate}

\paragraph{Training of the model}

The neural network models are trained for three epochs using the Adam optimization algorithm, minimizing a standard cross-entropy loss. 
We used a batch size of $2048$ and a learning rate of $\alpha = 0.001$. 
We initialized weights randomly from a Gaussian distribution with zero mean and $\sigma = 0.01$. 

\section{Data description}
\label{sec5}

\subsection{Overview}

An insurance company provided a data set consisting of claims from outpatient care. 
The data set comprises $381,013$ doctor's bills with $3.3$ million items in total. 
Each data point is a sequence of treatments encoded with anonymized IDs. 

There are $15$ input features in total.
In the data, we have two types of features for each patient: general and visit-specific.
General static features include age, sex, insurance type and doctor's speciality and refer to the patient, insurance, and doctor in general.
Visit-specific features describe each outpatient visit of a patient. As these features we use: treatments including type of a treatment, number of each treatment, cost of each treatment,  factor (multiplies the amount of a treatment because of potential complications), total amount of money charged,  billing type, cost category, and performance type. The type of treatment is coded using one of more than two thousand categories. 

\begin{figure}[ht!]
    \centering\includegraphics[width=300px]{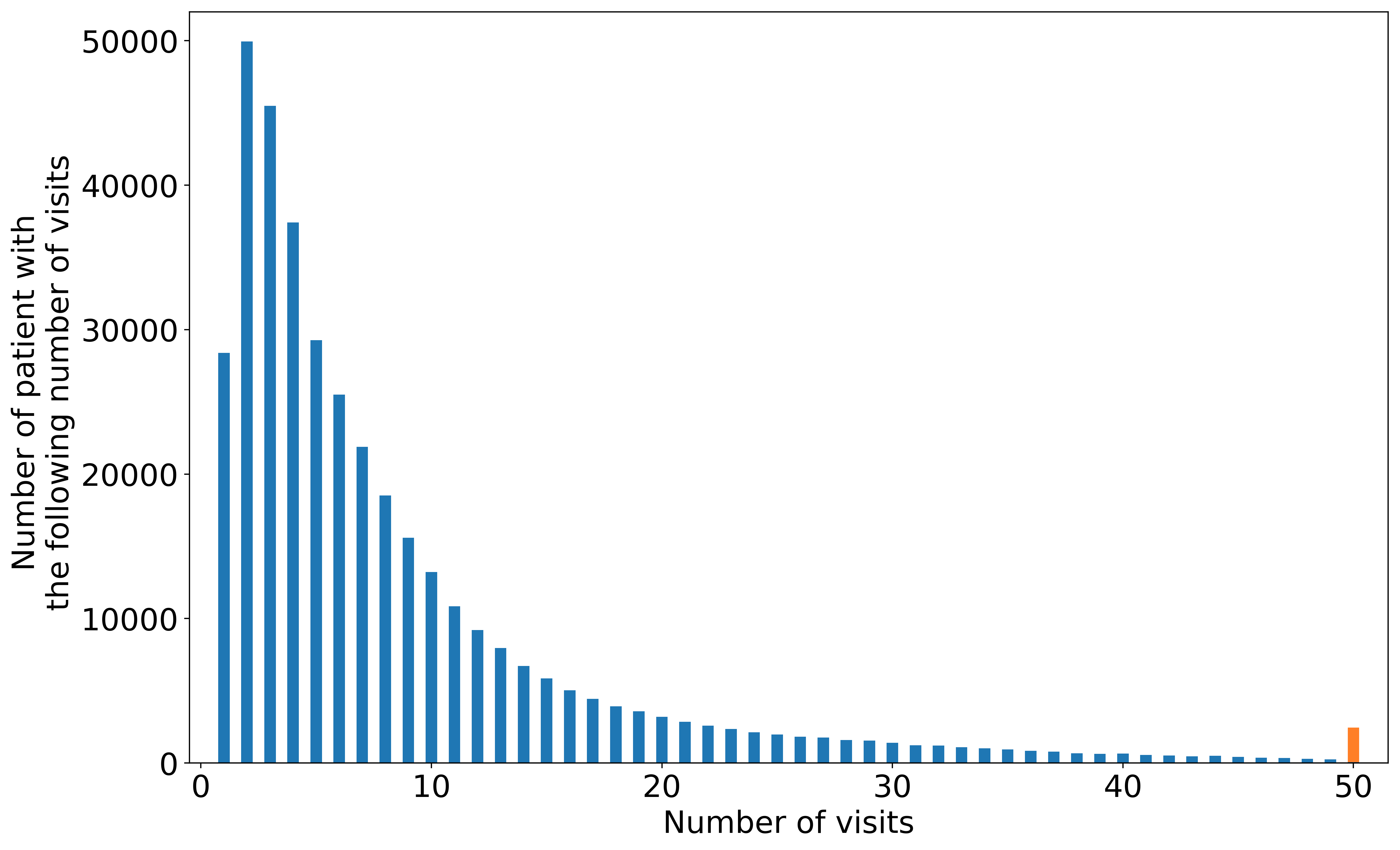}
    \caption{Histogram of the number of billing items for patients. The right orange bar represents the number of patients with $50$ or more. Most of the patients have fewer than $5$ treatments / items, and the most frequent number of billing items is $2$.}
    \label{fig:visits_freq}
\end{figure}

The number of treatments on a bill varies greatly among the patients.
Figure~\ref{fig:visits_freq} provides a  histogram of the number of treatments / items for the claims data. 
The distribution of treatments is nonuniform. Most patients have  only a small number of treatments in case of an outpatient treatment.

For each record we have a label.
The label is either ``fraudulent'' or ``non-fraudulent'' with fraudulent records corresponding to various fraudulent activities. Here ``fraudulent'' simply refers to the fact that the final amount of the bill was corrected, which can happen for different reasons.
About $2\%$ of records are fraudulent. 
The problem is to identify whether the record corresponds to a fraudulent activity based on given input features.

\subsection{Treatments}

The goal of our work is to determine whether information from labels of treatments can help to identify fraud in an automatic way. 
As the number of items / treatments varies for each patient, we must  aggregate information about all treatments in one vector: we want to construct an embedding of all treatments into a vector of a fixed dimensional size. An approach to deal with varying input size has been proposed by \cite{FLS}.

The natural way to construct embeddings is to use methods that have their roots in natural language processing,
as a doctor's bill is represented by a corresponding series of treatments.
Each anonymized treatment belongs to an alphabet of size $2205$. 
Treatments are summarized to $17$ upper-level groups. 
Another alphabet feature is the kind of benefit with $24$. 
There are $15$ categories for cost type in the data set.

\begin{figure}[ht!]
    \centering\includegraphics[width=200px]{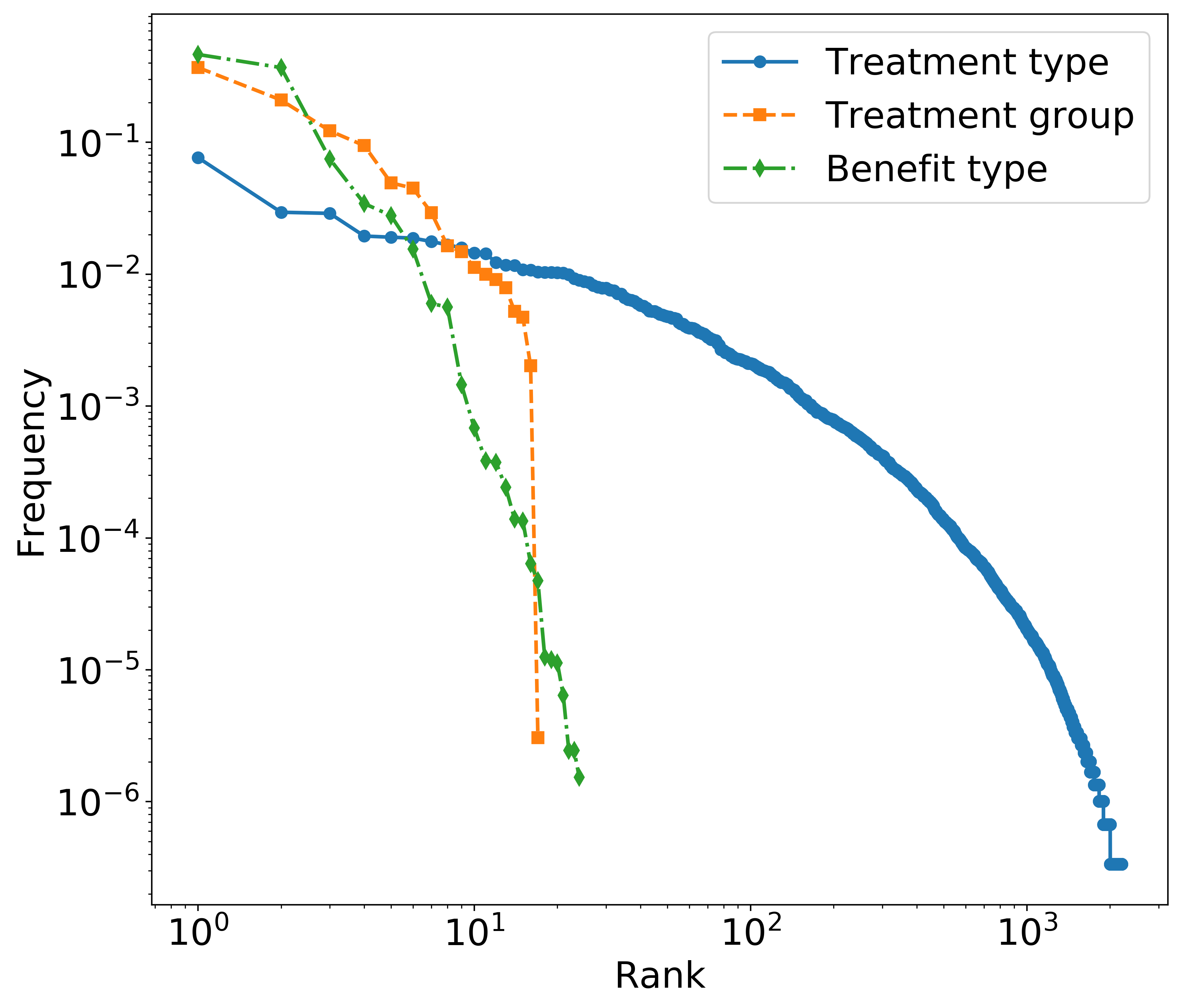}
    \caption{Log-log plot for ranks and corresponding frequencies of treatments, groups of treatments, and benefits in the data set. The plot significantly deviates from the straight line expected according to the Zipf's law}
    \label{fig:words_zipf}
\end{figure}

Moreover, the distribution of treatments with respect to their rank (e.g.,  the most frequent treatment has rank one) is close to what is known in natural language processing as empirical Zipf's law~\cite{montemurro2001beyond}, i.e. the frequency of any treatment is roughly inversely proportional to its rank in the frequency table.
Figure~\ref{fig:words_zipf} demonstrates this behavior for our data set as a log-log plot. 
However, we see a heavier tail with more rare treatments having higher frequencies than would be expected from the Zipf's law.
This means that there are fewer rare treatments compared to what happens in natural language texts.

Also note that a specific treatment does not correlate with fraud: if we measure correlation between the presence of a specific treatment and the target variable, the maximum absolute value of correlation is only $0.0243$.
We therefore need to apply more sophisticated machine learning approaches to make it possible to identify fraudulent treatments.

\section{Results}
\label{sec6}

\subsection{Metrics}

There are many metrics used to evaluate classification models. Because we have an imbalanced classification problem, the main goal is to detect the minority class with high precision. Thus, for fraud detection it is very important to find all actual positive events. When all instances of minority class are correctly predicted, then the solution is usually considered as excellent. 

Below we consider the common metrics for measuring the quality of imbalanced classification problems.

\begin{table}[h]
    \centering
    \begin{tabular}{c|cc}
         &Actual Positive& Actual Negative\\
        \hline
         Predicted Positive & True Positive (TP) & False Positive (FP)  \\
         Predicted Negative & False Negative (FN) & True Negative (TN) \\
    \end{tabular}
    \caption{Confusion Matrix for Binary Classification}
    \label{table:conf_matrix}
\end{table}

The confusion matrix is the basis of all metrics. From  Table \ref{table:conf_matrix}, commonly used metrics can be generated to estimate the performance of a classifier with different focuses of evaluation, such as the area under ROC curve (ROC AUC) and area under PR curve (PR AUC). These metrics are based on simpler metrics, such as recall (true positive rate), precision, and false positive rate. For this we introduce some notation.

\begin{enumerate}
    \item Recall = true positive rate $\frac{TP}{TP + FN}$ is the percentage of positive instances correctly classified. When this metric is equal to 1, it means that all fraud cases have been identified.
    \item Precision = $\frac{TP}{TP + FP}$ is the percentage of positive instances among positive predictions. A high value for precision means a good understanding of fraud behavior. 
    \item False Positive Rate = $\frac{FP}{FP + TN}$ is the percentage of positive instances misclassified.  
\end{enumerate}

The F1 score is defined as $2 \frac{\mathrm{Precision} \cdot \mathrm{Recall}}{\mathrm{Precision} + \mathrm{Recall}}$ and lies in the  interval $[0, 1]$. A hgiher F1 score is preferred.

The common ROC curve shows how well both classes are classified. The quality of one class is estimated by the high value of TPR, and the quality of the second class is estimated by the low value of FPR. Therefore, a good prediction is the ``balance'' between these values. 

PR curve characterizes how well the minor class (fraud class) is  classified. We want to maximize the TP number in predicted positive and actual positive classes. This curve is more appropriate for imbalanced data sets, because it reflects the model's ability to distinguish the fraud behavior from the common behaviour. Thus, the PR AUC metric is more focused on the minority class and, as a result, has an extra advantage compared to the other metrics, because it reflects the prediction quality of the most important class of the  problem. 

\subsection{Validation procedure}

To evaluate the performance of the model, we use data splitting. We randomly split up the dataset: $80\%$ of the bills are used during training, and the data from the remaining $20\%$ are used for testing of constructed models. 
Because the problem is imbalanced, we split the data into training and test data in a stratified way: ratios of classes in training and test samples coincide with those in the initial sample.

\subsection{Results}

\subsubsection{Usefulness of treatment features}

We have two types of features: general static features and visit-specific features.
In this section, we generate TF-IDF features from the visit-specific features and compare three different sets of features: only general, only visit-specific, and both.
The results are given in Table~\ref{tab:features_set_comparison}. The results clearly indicate that using all available features leads to the most accurate predictions, resulting in the highest ROC AUC and PR AUC value.

\begin{table}[h!]
    \centering
    \begin{tabular}{llll}
        \hline
        Features &  ROC AUC & PR AUC \\ 
        \hline
        General        & $0.8277$ & $0.0336$ \\ 
        Visit-specific & $0.8769$ & $0.1140$ \\ 
        Both           & $0.9126$ & $0.1516$ \\ 
        \hline
    \end{tabular}
    \caption{Quality of the model for three different sets of features: both general and visit-specific features improve the quality of the model}
    \label{tab:features_set_comparison}
\end{table}


\subsubsection{Overall performance of models}

We set gradient boosting hyperparameters --- the number of trees to $2000$ and the maximum depth to $10$. The embedding dimension for SWEM is $300$.
We estimate the performance metrics using 10-fold cross validation with $80\%$ data used for training and $20\%$ of the data used for the test each time.


The main results across the models are given in Table \ref{tab:results}. 
Embedding-based models outperform the gradient boosting classifier on the same data. 
For SWEM, the best aggregation strategy is max-pooling.
Adding of general features improves the quality of both gradient-boosting-based and neural-network-based models.

\begin{table}[h!]
\centering
\begin{tabular}{lccc}
\hline
\textbf{Model}                  & \textbf{Static} & \textbf{ROC AUC} & \textbf{PR AUC} \\
&\textbf{features} && \\ \hline
Gradient boosting (BoW)                   & without                                    & $ 0.8625  \pm  0.0297  $                                 & $ 0.1806  \pm  0.0446  $                                    \\
Gradient boosting (TF-IDF)                & without                                    & $ 0.8625  \pm  0.0297  $                                 & $ 0.1928  \pm  0.0658  $                                     \\ 
Gradient boosting (BoW)                   & with                                     & $ 0.8934  \pm  0.0281  $                                 & $ 0.1958  \pm  0.0579  $                                     \\

Gradient boosting (TF-IDF)                & with                                     & $ 0.8948  \pm  0.0303  $                                & $ 0.2036  \pm  0.0659  $                                    \\
\hline
SWEM-mean & without                                    &   $ 0.8753  \pm  0.0233 $                               &   $ 0.2052  \pm  0.0778 $                                       \\
SWEM-concat & without                                    &   $ 0.8551  \pm  0.0302 $                               &   $ 0.1782  \pm  0.0754 $                                       \\
SWEM-max & without                                    & $ 0.8957  \pm  0.0217 $                                 &    $ 0.2279 \pm  0.0677 $                                    \\
SWEM-mean & with                                    &   $ 0.8932  \pm  0.0321 $                             &   $ 0.2112  \pm  0.0804 $                                     \\
SWEM-concat & with                                    &   $ 0.8696  \pm  0.0377 $                            &   $ 0.1784  \pm  0.0814 $                             \\
SWEM-max & with                                    &   $ \mathbf{0.9062  \pm  0.0252} $                             &   $ \mathbf{0.2445  \pm  0.0867} $                             \\
\hline
\end{tabular}
\caption{\label{tab:results} Quality of used models for $10$-fold cross validation: mean values and standard deviations given after $\pm$ sign.  SWEM model with different aggregation strategies mean, concat, and max models that train problem-specific embeddings perform better than gradient boosting with various types of features. Adding general static features significantly reduces quality of the neural network model. Embedding dimension for SWEM is equal to $300$.}
\end{table}


\subsection{Dependence of quality of models on sample size}

\begin{figure}
    \centering
    \begin{subfigure}[b]{0.43\textwidth}
        \includegraphics[width=\textwidth]{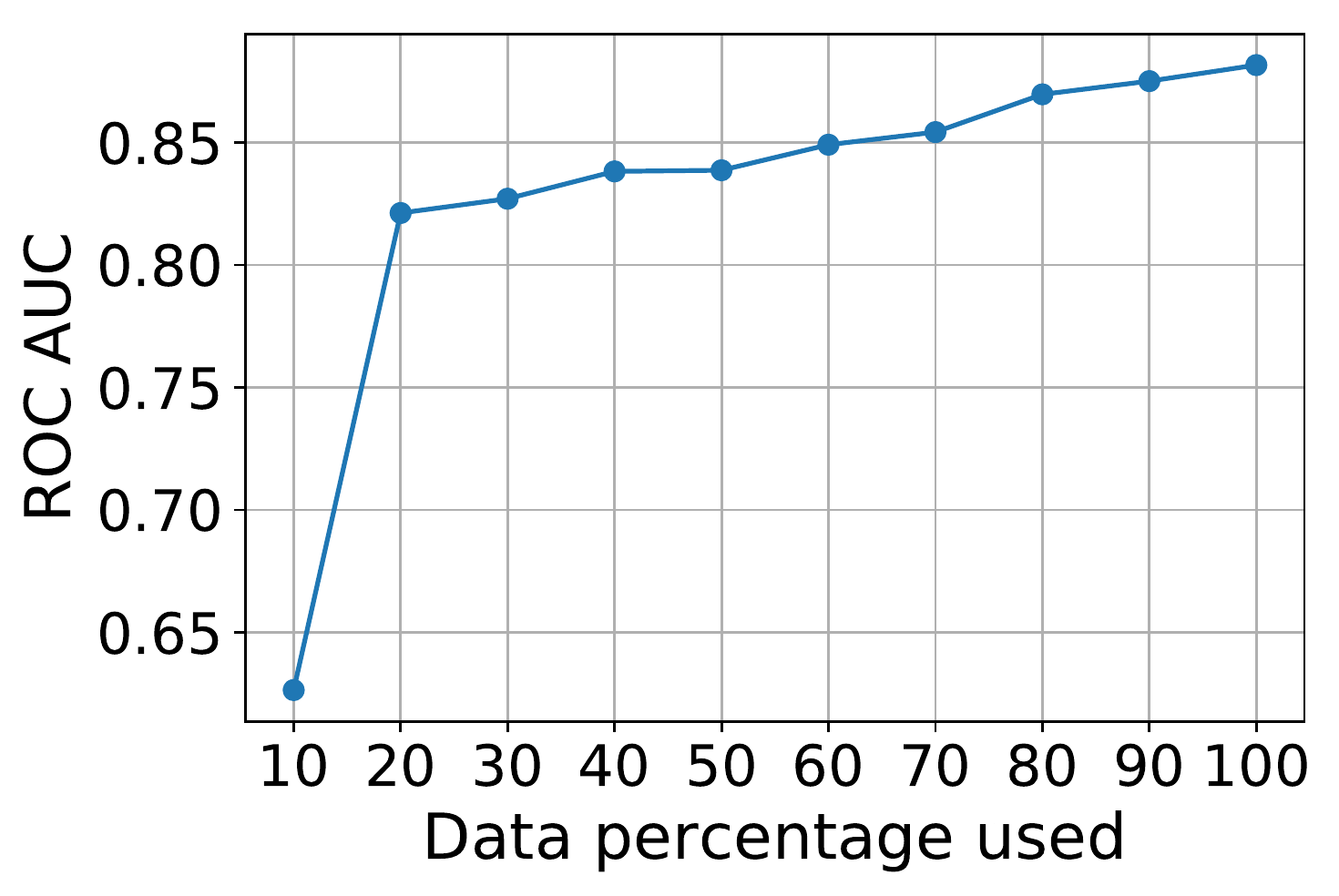}
        \caption{Dependence of ROC AUC on training data size}
    \end{subfigure}
    ~
    \begin{subfigure}[b]{0.43\textwidth}
        \includegraphics[width=\textwidth]{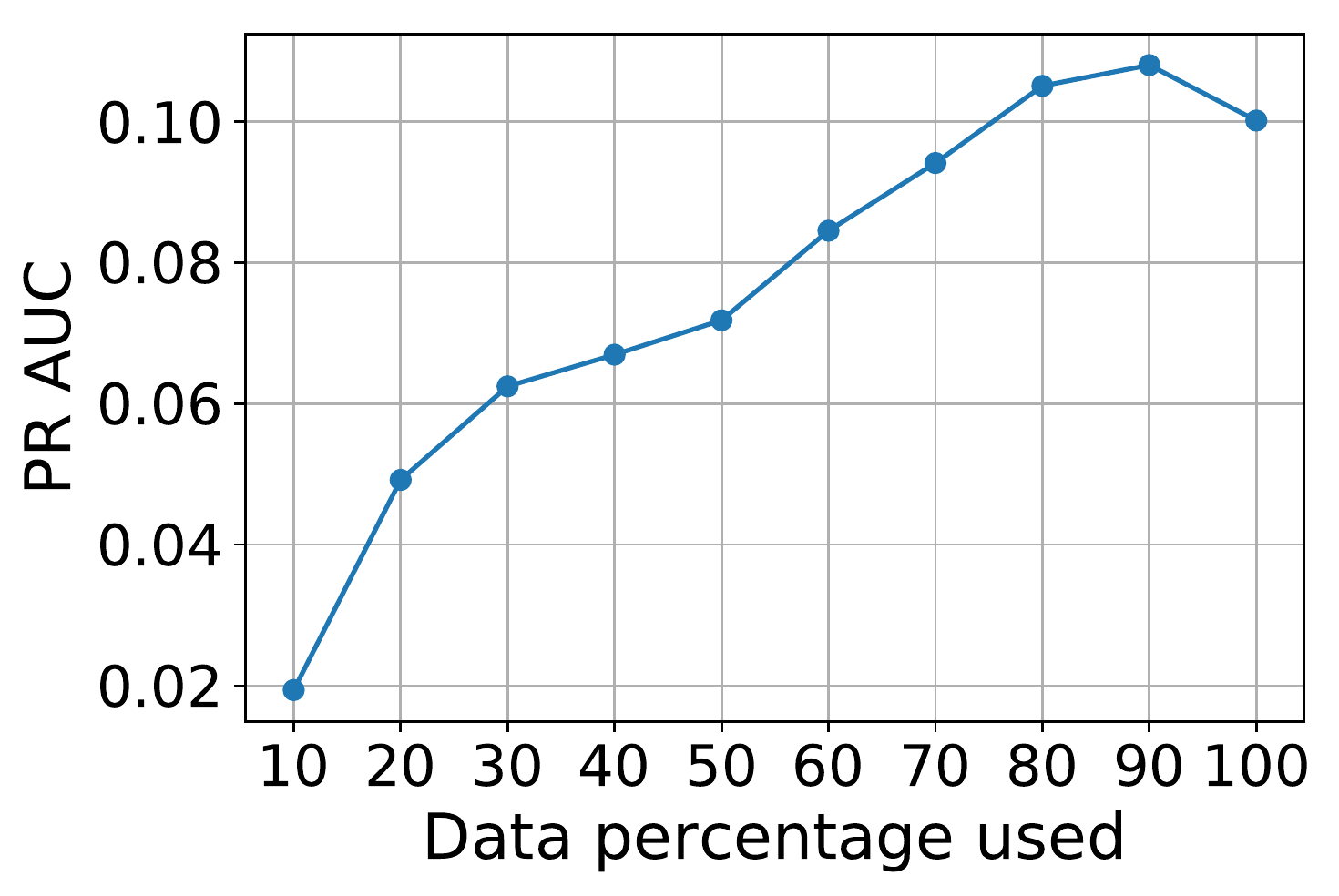}
        \caption{Dependence of PR AUC on training data size}
    \end{subfigure}
    \caption{Dependence of model quality on used sample size. Increasing the proportion of training data leads to further increase in the  quality of the fraud detection model. Results are provided for the SWEM-max based model}
    \label{fig:quality_size}
\end{figure}

We examine how the proportion of the data used for training affects the  quality of the final model.
The model was trained using $10$, $20$, ..., $100$ percent of the initial training sample selected at random.
The results are given in Figures~\ref{fig:quality_size}. We see that the PR AUC and ROC AUC still increase as we increase the proportion of training data passed to the model.

\subsubsection{Selection of NN architecture hyperparameters}

\begin{figure}
    \centering
    \begin{subfigure}[b]{0.43\textwidth}
        \includegraphics[width=\textwidth]{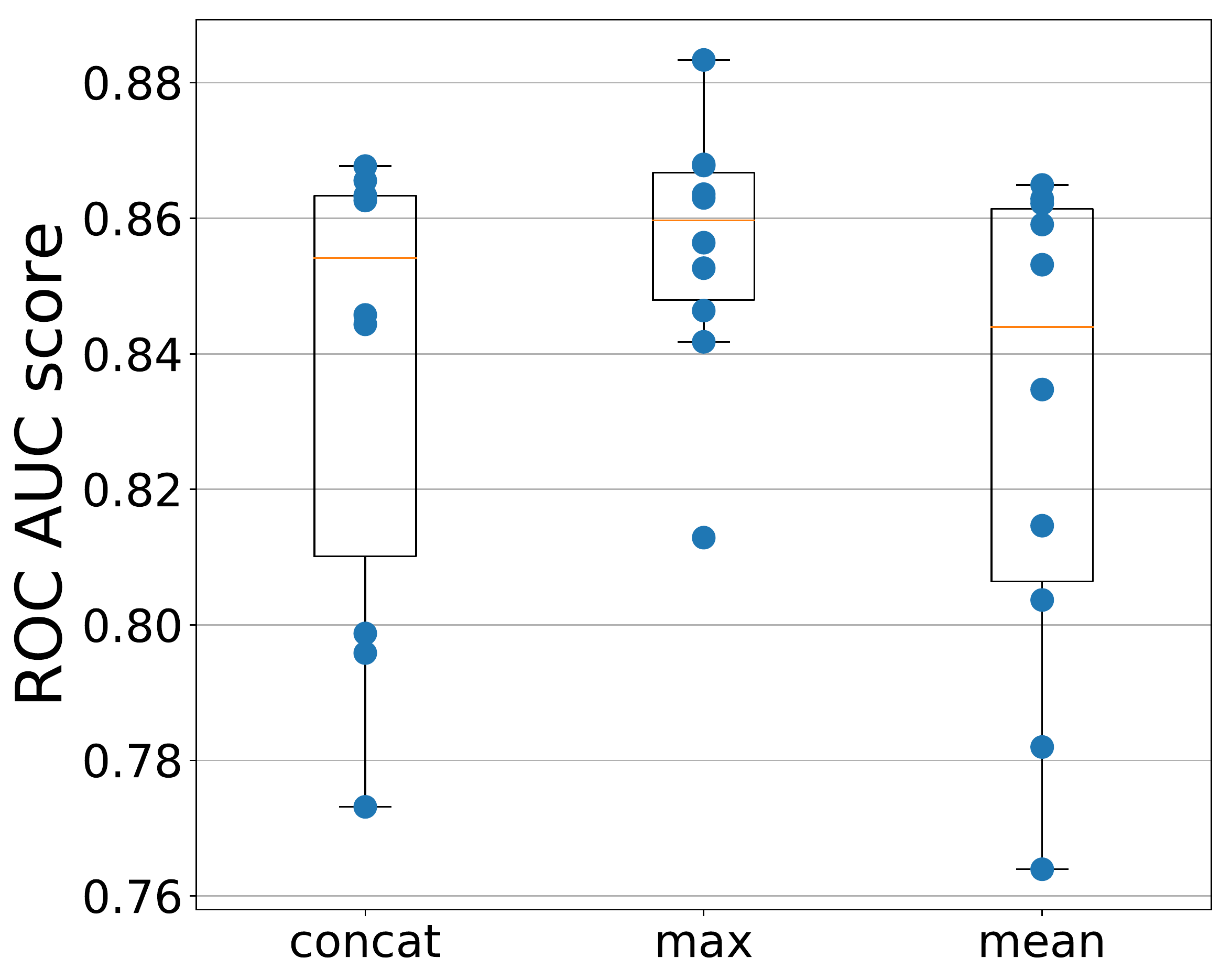}
        \caption{Dependence of ROC AUC on a strategy for aggregating  embeddings of treatments to an embedding of a patient}
    \end{subfigure}
    ~ 
    \begin{subfigure}[b]{0.43\textwidth}
        \includegraphics[width=\textwidth]{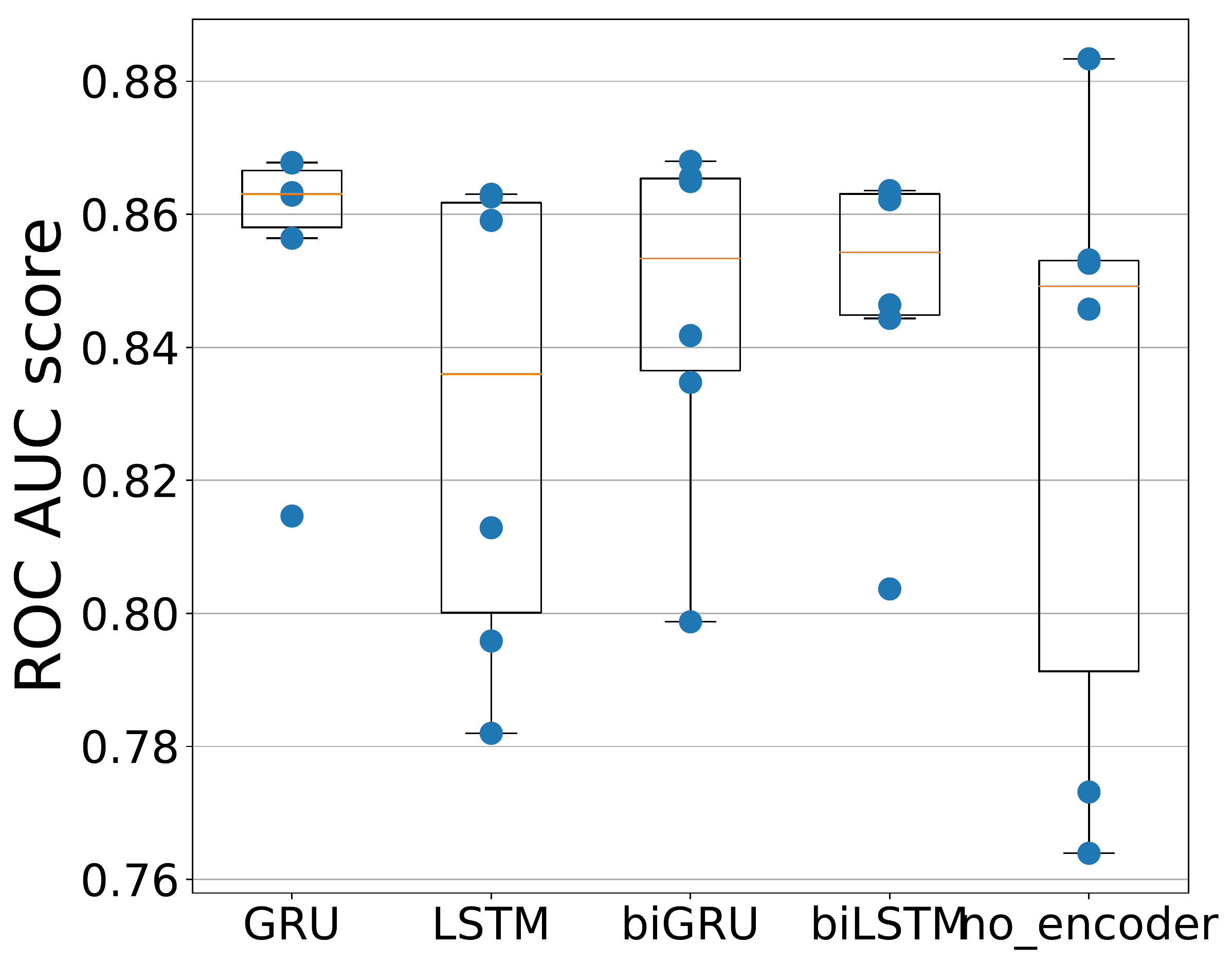}
        \caption{Dependence of ROC AUC on a strategy for encoding  embeddings  $\phantom{aaaaaa}$
        $\phantom{aaaaaa}$}
    \end{subfigure}
    ~ 
    \begin{subfigure}[b]{0.43\textwidth}
        \includegraphics[width=\textwidth]{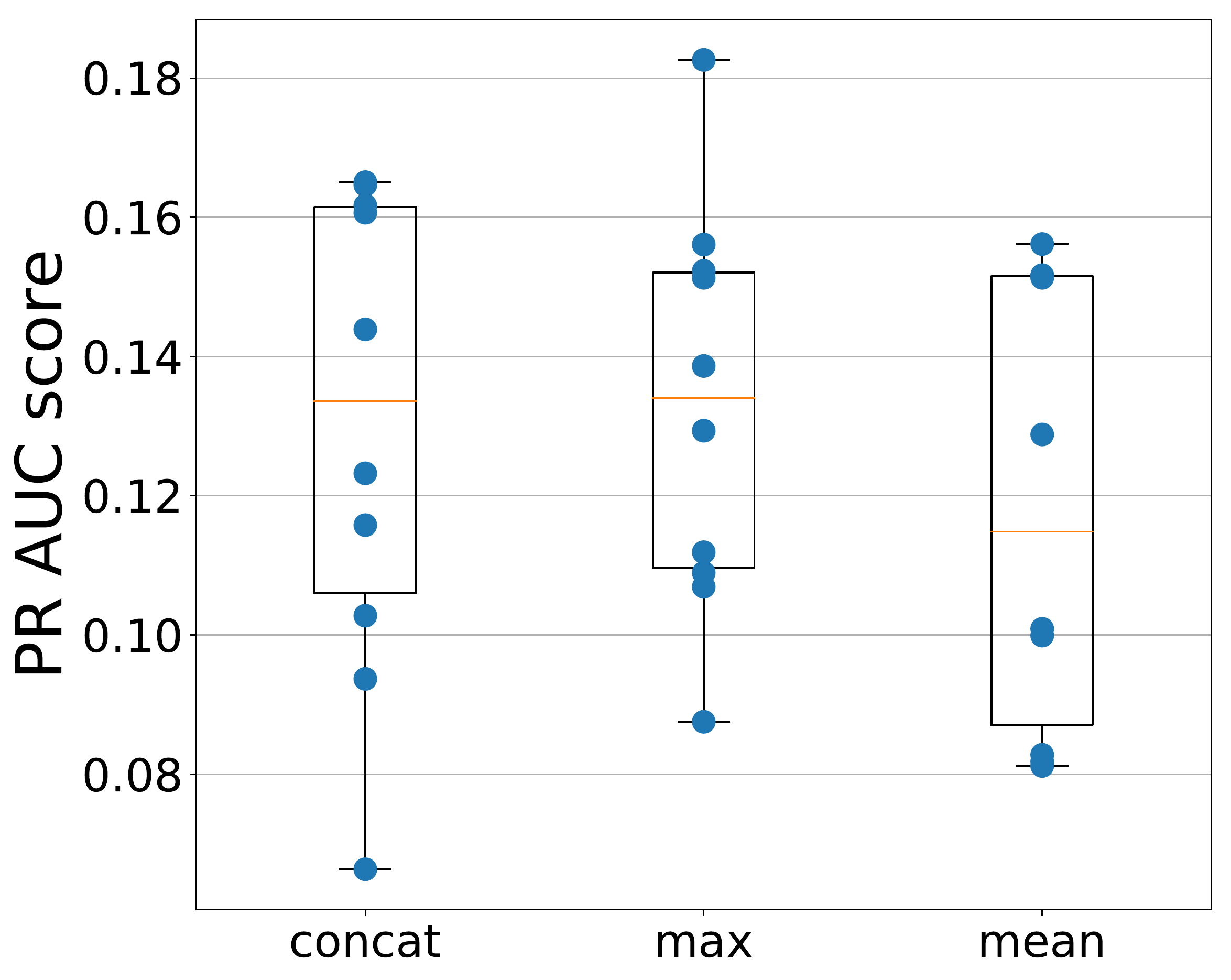}
        \caption{Dependence of PR AUC on a strategy for aggregating  embeddings of treatments to an embedding of a patient}
        \label{fig:pr_auc_aggregation}
    \end{subfigure}
    ~ 
    \begin{subfigure}[b]{0.43\textwidth}
        \includegraphics[width=\textwidth]{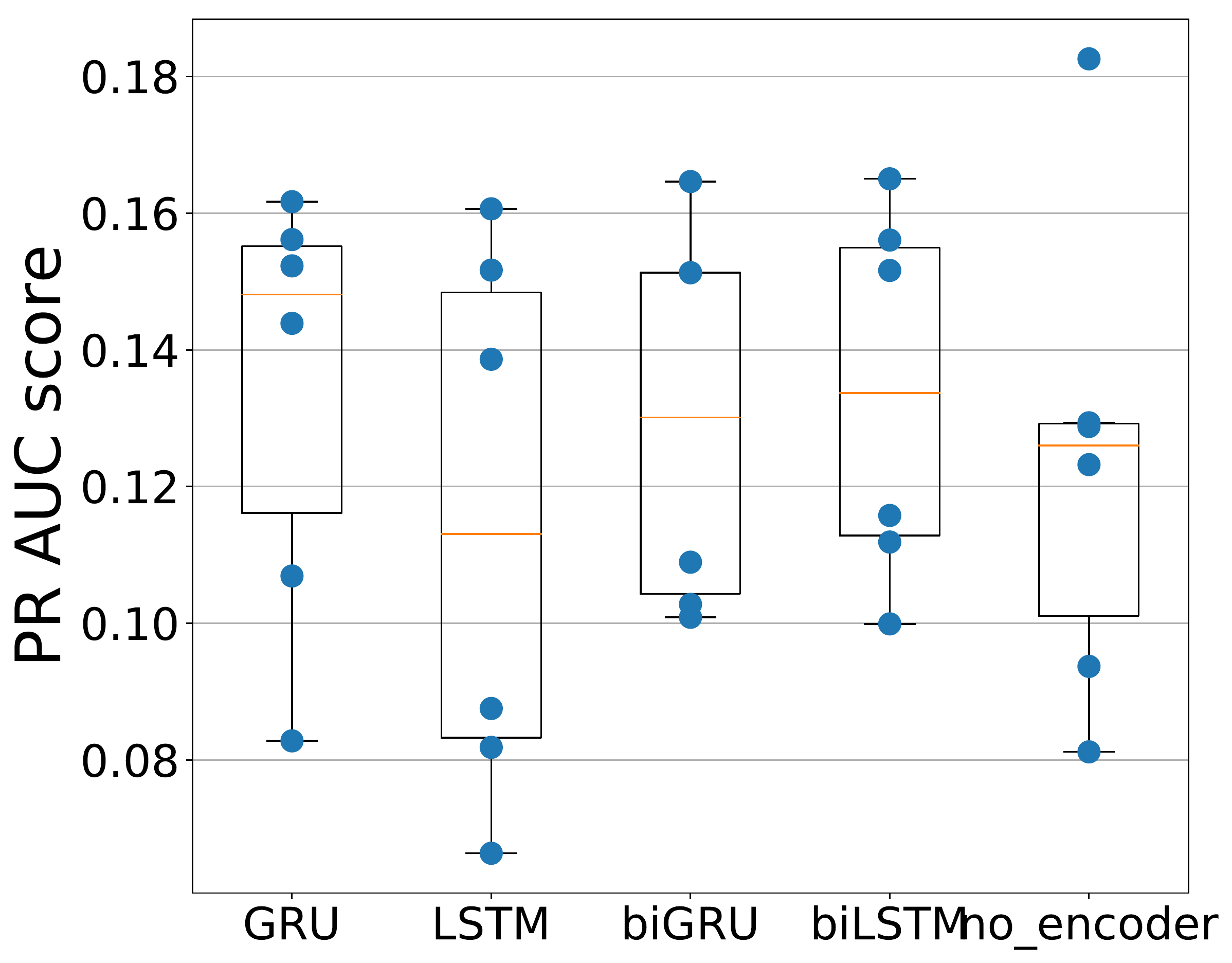}
        \caption{Dependence of PR AUC on a strategy to construct encodings for embeddings $\phantom{aaaaaaaaaaaaa}$}
    \end{subfigure}
    \caption{Dependence of the  model quality on the model architecture: selection of the right encoding strategy and the right aggregation strategy lead to an improvement in the quality of the model}
    \label{fig:hyper_selection}
\end{figure}

We undertake cross-validation to test how the selection of hyperparameters affects the performance of the model.
In particular we consider: inclusion and exclusion of general features in our neural network model, use of different encoders (see the description of the Embedding Layer in Section \ref{aggr}) in the architecture of the neural network, and different types of aggregation of treatment embeddings to get an embedding, characterizing a particular patient.
The results are given in Figure~\ref{fig:hyper_selection}.

\subsubsection{Selection of embedding dimension}

\begin{figure}
    \centering
    \begin{subfigure}[b]{0.43\textwidth}
        \includegraphics[width=\textwidth]{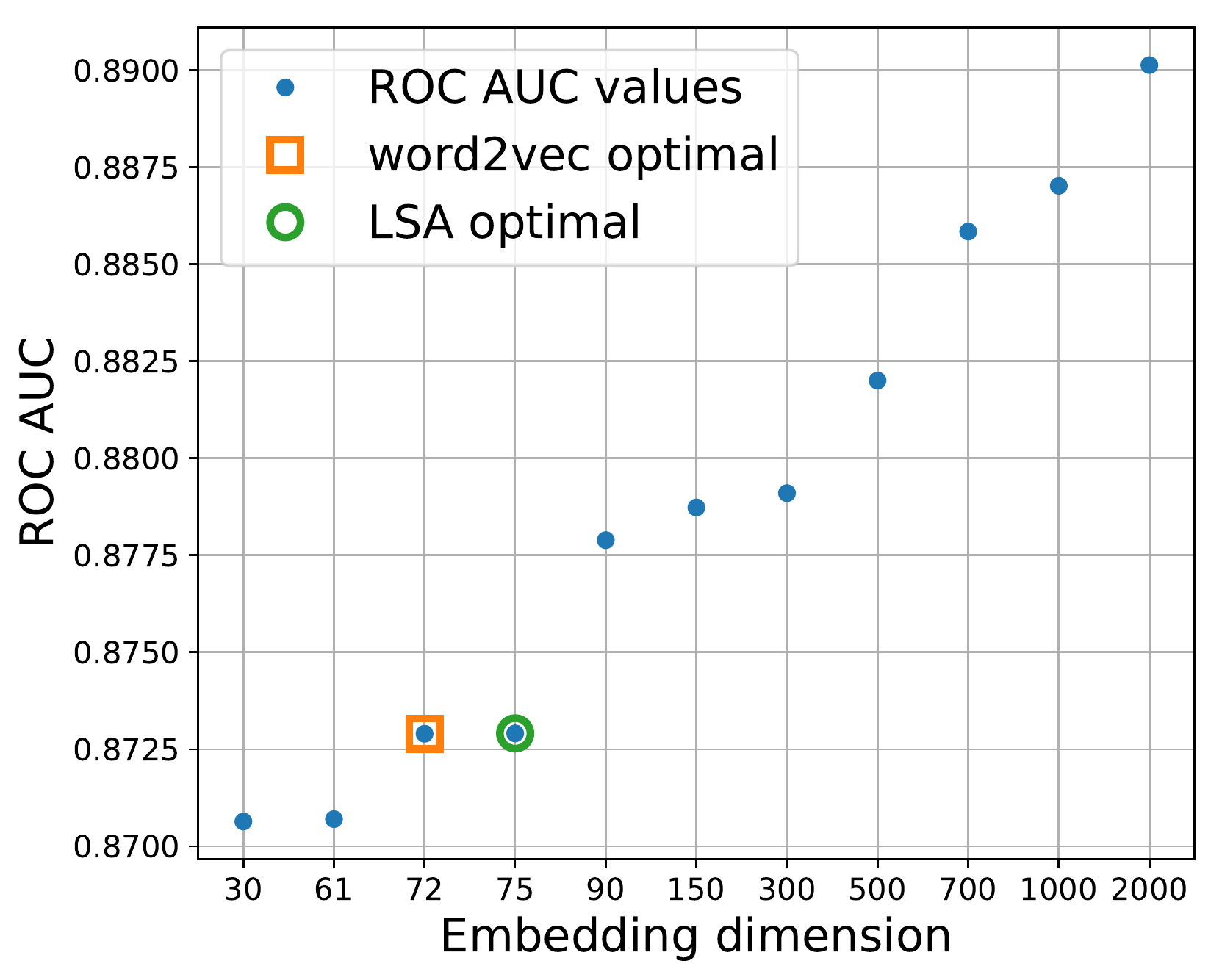}
        \caption{Dependence of ROC AUC on dimension of embedding}
        \label{fig:roc_auc_d}
    \end{subfigure}
    ~ 
    \begin{subfigure}[b]{0.43\textwidth}
        \includegraphics[width=\textwidth]{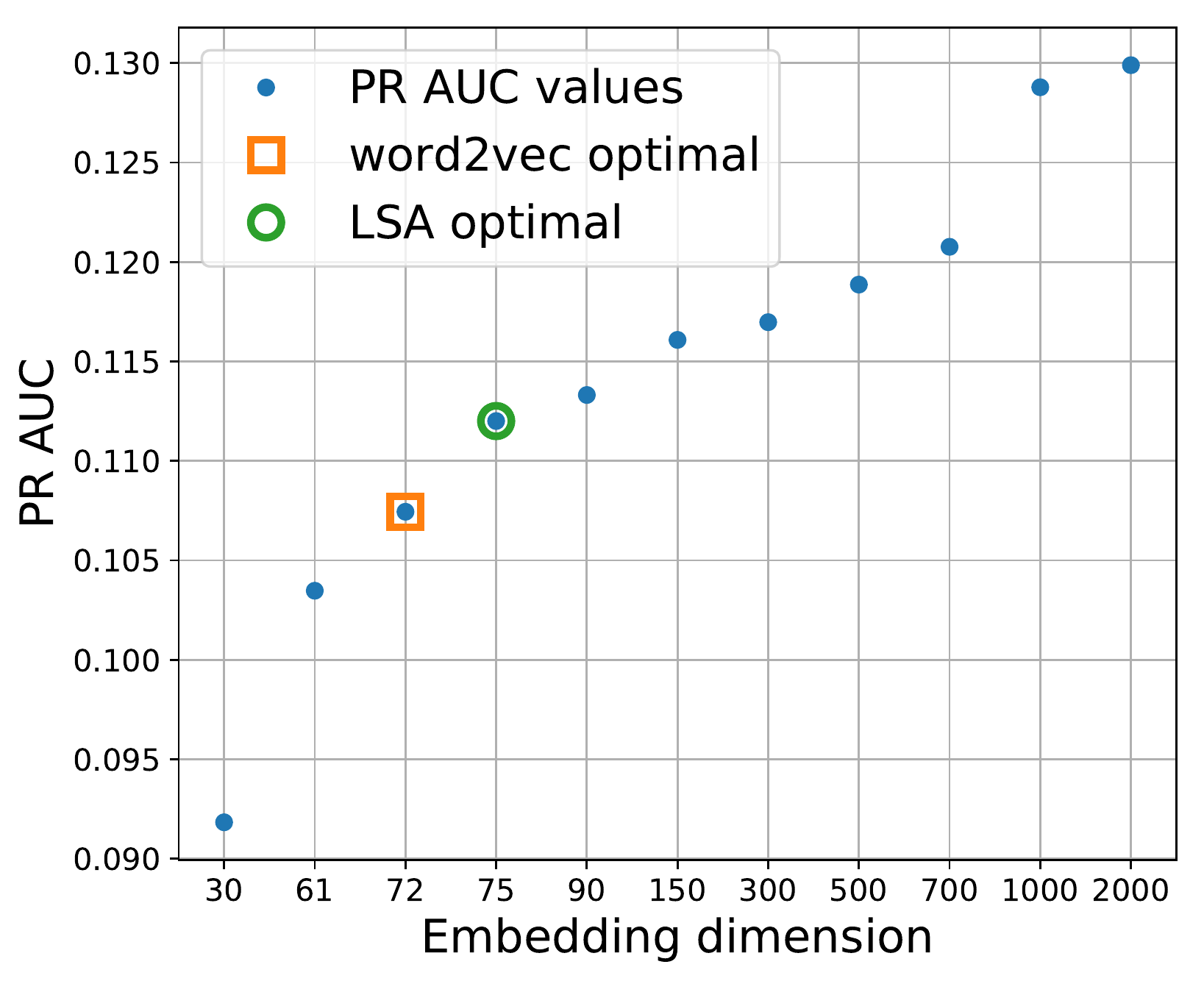}
        \caption{Dependence of PR AUC on dimension of embedding}
        \label{fig:pr_auc_d}
    \end{subfigure}
    \caption{Dependence of model quality on dimension $d$ of embeddings: increasing of $d$ leads to better quality of models. To estimate the optimal dimension $d$ we use the state-of-the-art methods by ~\cite{yin2018dimensionality}. These dimensions yield lower quality of models.
    The scale of the plots is not uniform.}
    \label{fig:embedding_dimension}
\end{figure}

To select the embedding dimension, we try models with different sizes of embedding dimension $d$.
Figure~\ref{fig:embedding_dimension} demonstrates that the quality of the  model gradually rises as the embedding dimension increases if we train embeddings along with the model parameters.
Our hypothesis is that direct use of natural language processing approaches  leads to loss of some information during mapping from the initial feature space to an embedded space,  resulting in poorer model performance. That is why increasing the embedding dimension lead to increase in accuracy, as we loose less information.

\subsubsection{Application of resampling}

As we are dealing with fraud detection data, the number of good cases is significantly higher than number of fraudulent cases. 
There are a number of approaches on imbalanced classification described in Section~\ref{sec:imb_classificaion}.
Most of the problems consider resampling techniques: how should we change the balance of classes in our training sample, for example by  dropping some major class objects or giving more weight to minor class objects.

In Table~\ref{tab:imbalanced} we provide results on how imbalanced classification approaches can improve the overall quality of the model.
We see that it is really the case for all applied resampling approaches.
The best techniques for the problem at hand are 
under-sampling (InstanceHardnessThreshold) and a 
combination of over- and under-sampling (SMOTEENN), see \cite{pacpacpac}.

\begin{table}[h!]
    \centering
    \begin{tabular}{lll}
    \hline
        Resampling technique & ROC AUC & PR AUC \\
         \hline
        No resampling (baseline) & $0.8345$ & $0.0994$ \\
        Over (SMOTE) & $0.844$ & $0.0961$ \\
        Over (ADASYN) & $0.8455$ & $0.0985$ \\
        Under (RepeatedEditedNN) & $0.8439$ & $0.1019$ \\
        Under (InstanceHardnessThreshold) & $\mathbf{0.8515}$ & $0.1011$ \\
        Both over- and under (SMOTEENN) & $0.8485$ & $\mathbf{0.1069}$ \\
         \hline
    \end{tabular}
    \caption{Improvement of gradient boosting model using resampling techniques for fighting imbalanced classification problem at hand: over- and under-sampling techniques are considered}
    \label{tab:imbalanced}
\end{table}

\subsubsection{Reliability of models}

Significant issues for machine learning models are reliability and resistance to malicious attacks.
In particular, this problem is important in fraud detection: if a malicious user of a decision system can provide slightly distorted data to the system and trick it, then the system is of limited use, see ~\cite{papernot2017practical}.

Usually one considers two approaches:
``is a model robust with respect to random errors in data submitted to a system?'' (~\cite{saltelli2004sensitivity}), and ``is a model robust to malicious efforts, when someone tries to break the system in a particular way by corrupting the input to the system?'' (~\cite{papernot2017practical}).

In our case we test the reliability with respect to the history of treatments: can we change it slightly and thereby obtain an entirely  different outcome with the model?
To do so, we compare the  quality the model before and after the corruption of the data.

We test these issues in two ways:
\begin{itemize}
    \item To test the reliability of the model, we randomly add a treatment to the end of the sequence of treatments for each customer. This disturbance corrupts the labels obtained by our model, as we use a different set of features.
    \item To test the model's resistance to malicious attacks, we randomly add $100$ treatments one by one to fraudulent cases and maliciously select model output that is affected the most by a single treatment addition. 
\end{itemize}
We use the machine learning model 
that was constructed using TF-IDF features for treatments and general features.

\begin{figure}
    \centering
    \begin{subfigure}[b]{0.43\textwidth}
        \includegraphics[width=\textwidth]{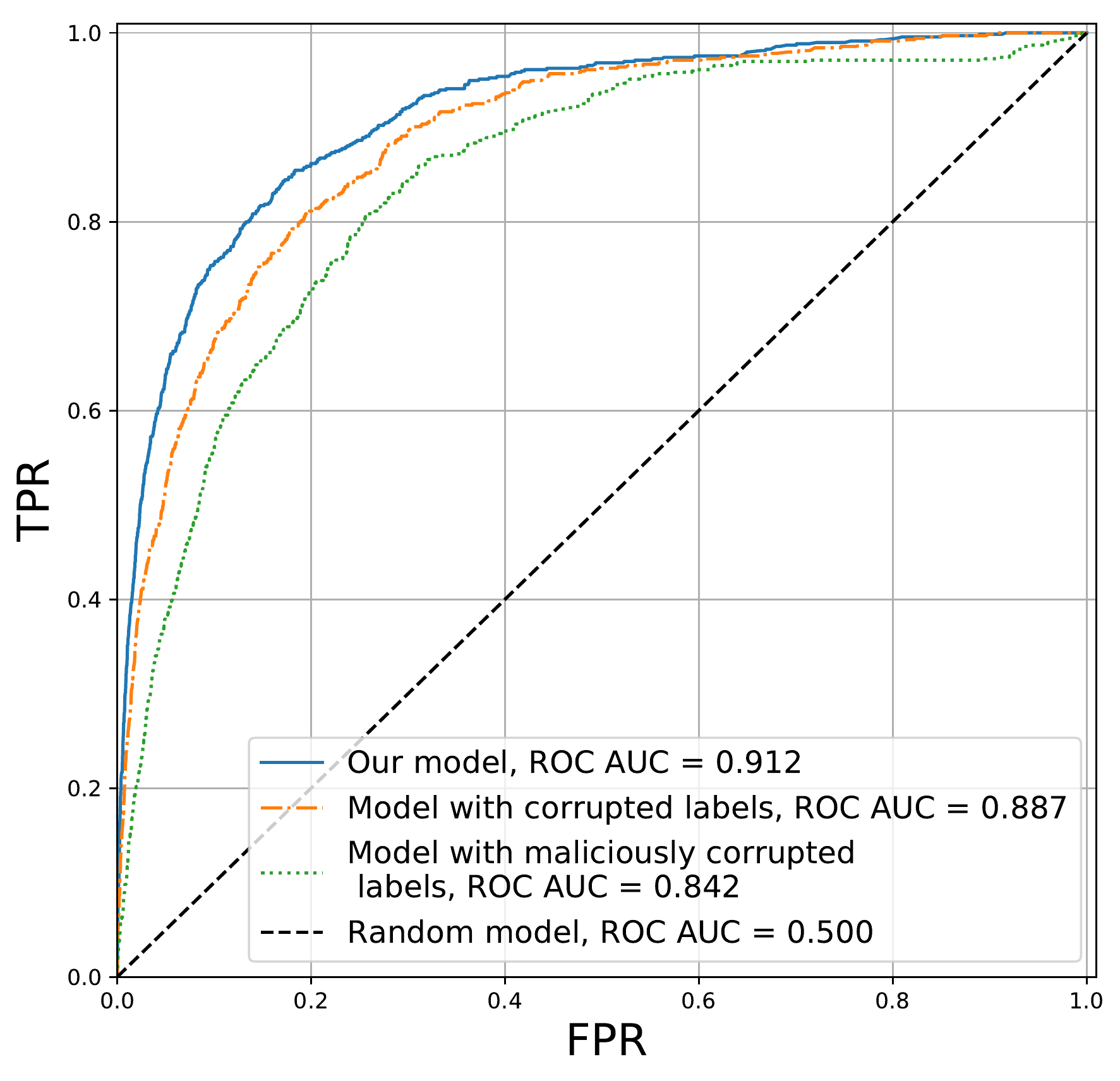}
    \end{subfigure}
    ~ 
    \begin{subfigure}[b]{0.43\textwidth}
        \includegraphics[width=\textwidth]{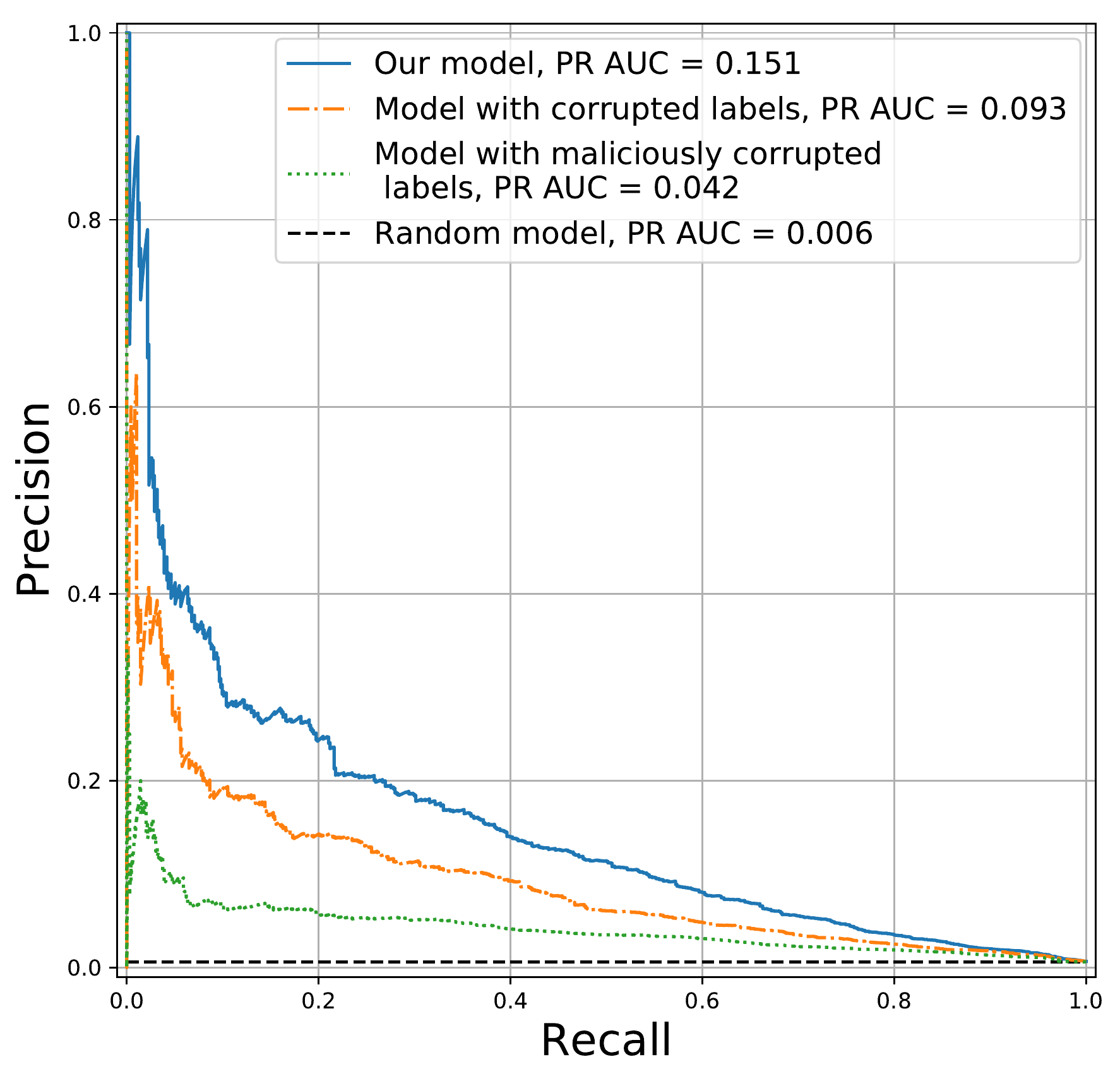}
    \end{subfigure}
    \caption{Comparison of ROC and PR AUC curves before and after corruption of inputs of the ``TF-IDF + Gradient Boosting'' model for the test data. Random addition of a treatment results in a small drop in the quality. Malicious attack results in a significant drop in the quality of the model.}
    \label{fig:corrupted_labels}
\end{figure}

\begin{figure}
    \centering
    \begin{subfigure}[b]{0.43\textwidth}
        \includegraphics[width=\textwidth]{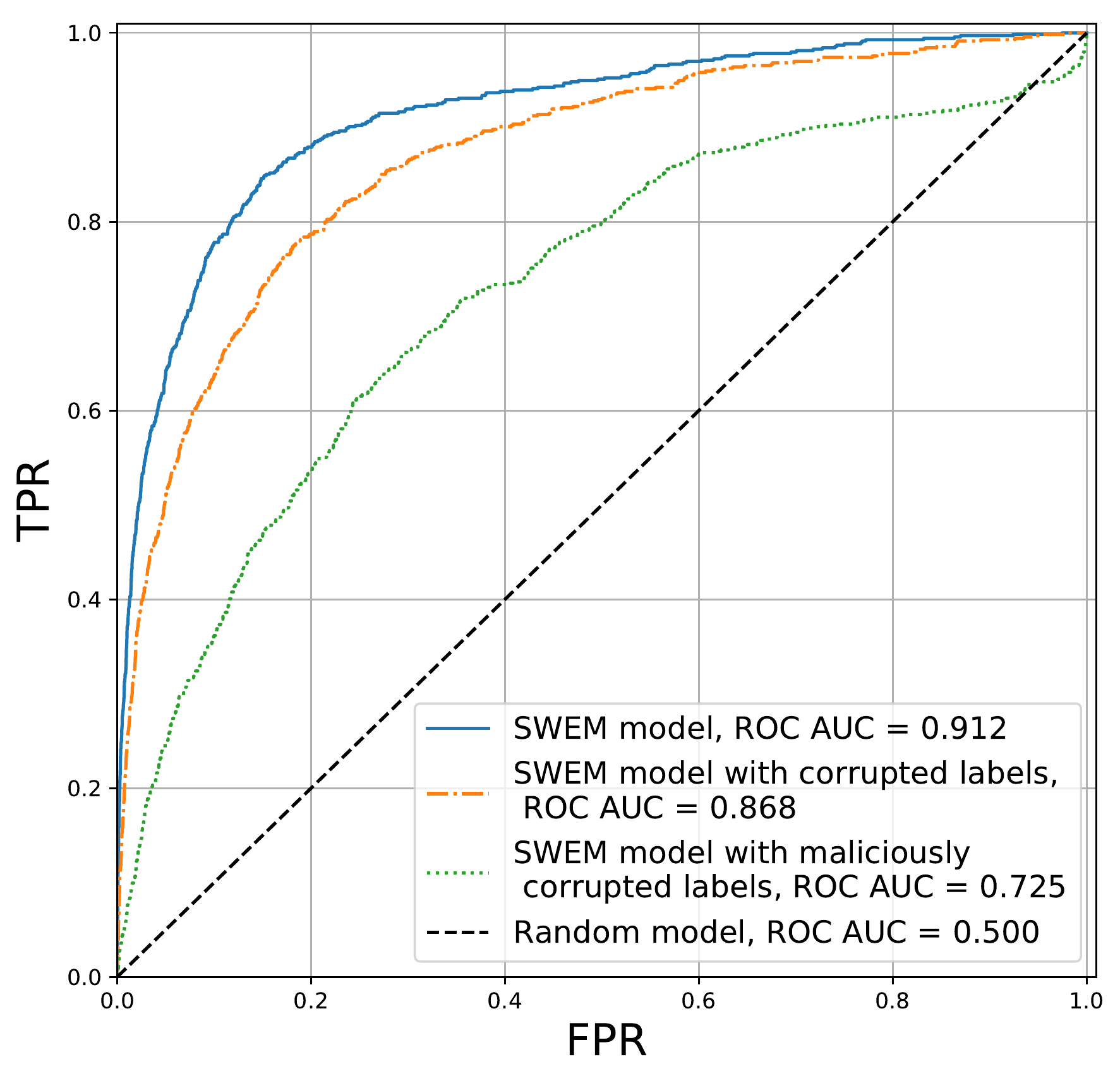}
    \end{subfigure}
    ~ 
    \begin{subfigure}[b]{0.43\textwidth}
        \includegraphics[width=\textwidth]{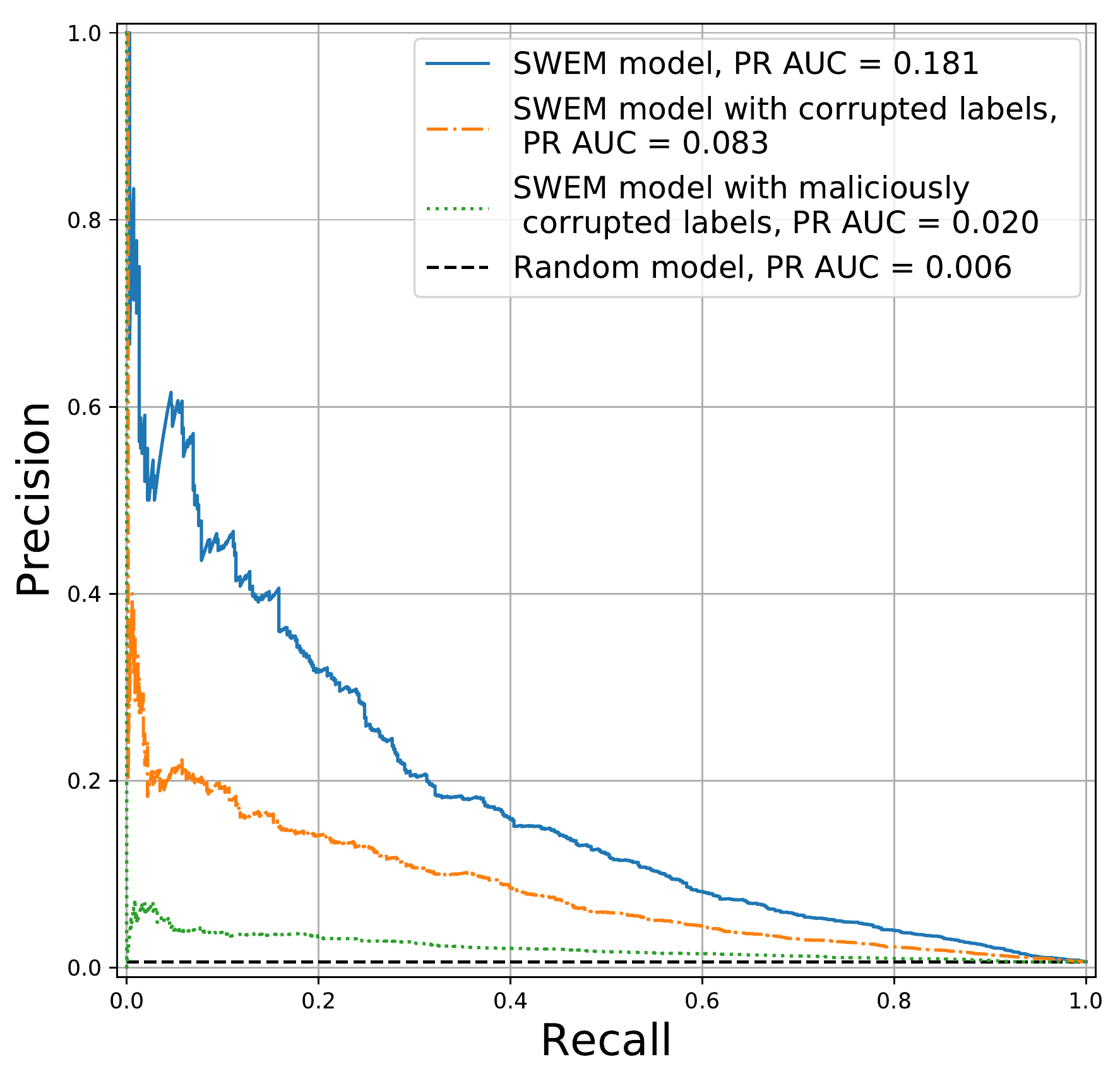}
    \end{subfigure}
    \caption{Comparison of ROC and PR AUC curves before and after corruption of inputs of the SWEM-based model for the test data. Random addition of a treatment results in a small drop in the quality. Malicious attack results in a significant drop in the quality of the model.}
    \label{fig:corrupted_labels_swem}
\end{figure}

The ROC and PR curves before and after the corruption of labels are shown in Figure~\ref{fig:corrupted_labels} for TF-IDF and gradient boosting approach and in Figure~\ref{fig:corrupted_labels_swem} for the SWEM-max approach.
We see that changing of inputs to the model leads to a drop in its quality  especially of the SWEM model.
To make the model more stable one should augment the training data with more cases and possible distortions of the initial data (so-called data augmentation) and keep the model secret to avoid a malicious attack.

\section{Conclusion}
\label{sec7}

In this paper we introduce and propose deep learning architectures that are tailor-made for claims data and provide embeddings for treatment codes / classification of diseases. Insurance fraud is one of the main threats to insurance and other financial companies. 
Using a real world data set, we show that our proposed methods based on embeddings for unstructured data outperform standard methods and have the  potential to improve the claims management process. Although doctor's bills have a text format, claims data and treatment codes seem to differ from ``traditional'' texts. Compared to other texts, in that they are more uniformly distributed among patients and the optimal dimension of embeddings appear to be higher as proposed for texts. By designing a tailor-made architecture for treatment embeddings we outperform standard models.
Our empirical experiments show that our model can be further improved by optimization of the neural network architecture and by increasing volume of data used for training. Moreover, our model is robust, to some extent,  to disturbances of the data and adversarial and malicious changes to some extent.

As digitization continues to proliferate, increasing amounts of  unstructured data in the form of texts will become available, including electronic health records and claims data,  personnel files and financial statements. Often these data will have a special structure and in particular  will contain variables with a large number of categories that  cannot be handled by classical econometric methods. The deep learning architectures and embeddings we propose in this paper are tailor-made for such data such as these and can be useful to researchers in health economics, organizational economics, finance, and many other fields.




%
%
%


\bibliographystyle{elsarticle-num-names} %
\bibliography{references}

\begin{thebibliography}{57}
\expandafter\ifx\csname natexlab\endcsname\relax\def\natexlab#1{#1}\fi
\providecommand{\url}[1]{\texttt{#1}}
\providecommand{\href}[2]{#2}
\providecommand{\path}[1]{#1}
\providecommand{\DOIprefix}{doi:}
\providecommand{\ArXivprefix}{arXiv:}
\providecommand{\URLprefix}{URL: }
\providecommand{\Pubmedprefix}{pmid:}
\providecommand{\doi}[1]{\href{http://dx.doi.org/#1}{\path{#1}}}
\providecommand{\Pubmed}[1]{\href{pmid:#1}{\path{#1}}}
\providecommand{\bibinfo}[2]{#2}
\ifx\xfnm\relax \def\xfnm[#1]{\unskip,\space#1}\fi
\bibitem[{Artemov et~al.(2015)Artemov, Burnaev, and Lokot}]{QuasiPeriodic}
\bibinfo{author}{A.~Artemov}, \bibinfo{author}{E.~Burnaev},
  \bibinfo{author}{A.~Lokot},
\newblock \bibinfo{title}{Nonparametric decomposition of quasi-periodic time
  series for change-point detection},
\newblock in: \bibinfo{booktitle}{Proc. SPIE}, volume \bibinfo{volume}{9875},
  \bibinfo{year}{2015}, pp. \bibinfo{pages}{9875 -- 9875 -- 5}.
  \DOIprefix\doi{10.1117/12.2228370}.
\bibitem[{Ishimtsev et~al.(2017)Ishimtsev, Bernstein, Burnaev, and
  Nazarov}]{kNN2017}
\bibinfo{author}{V.~Ishimtsev}, \bibinfo{author}{A.~Bernstein},
  \bibinfo{author}{E.~Burnaev}, \bibinfo{author}{I.~Nazarov},
\newblock \bibinfo{title}{Conformal k-nn anomaly detector for univariate data
  streams},
\newblock in: \bibinfo{editor}{A.~Gammerman}, \bibinfo{editor}{V.~Vovk},
  \bibinfo{editor}{Z.~Luo}, \bibinfo{editor}{H.~Papadopoulos} (Eds.),
  \bibinfo{booktitle}{Proceedings of the Sixth Workshop on Conformal and
  Probabilistic Prediction and Applications}, volume~\bibinfo{volume}{60} of
  \textit{\bibinfo{series}{Proceedings of Machine Learning Research}},
  \bibinfo{publisher}{PMLR}, \bibinfo{address}{Stockholm, Sweden},
  \bibinfo{year}{2017}, pp. \bibinfo{pages}{213--227}.
\bibitem[{Artemov and Burnaev(2016)}]{Degradation2016}
\bibinfo{author}{A.~Artemov}, \bibinfo{author}{E.~Burnaev},
\newblock \bibinfo{title}{Detecting performance degradation of
  software-intensive systems in the presence of trends and long-range
  dependence},
\newblock in: \bibinfo{booktitle}{2016 IEEE 16th International Conference on
  Data Mining Workshops (ICDMW)}, \bibinfo{year}{2016}, pp.
  \bibinfo{pages}{29--36}. \DOIprefix\doi{10.1109/ICDMW.2016.0013}.
\bibitem[{Smolyakov et~al.(2018)Smolyakov, Sviridenko, Burikov, and
  Burnaev}]{OCSVM2018}
\bibinfo{author}{D.~Smolyakov}, \bibinfo{author}{N.~Sviridenko},
  \bibinfo{author}{E.~Burikov}, \bibinfo{author}{E.~Burnaev},
\newblock \bibinfo{title}{Anomaly pattern recognition with privileged
  information for sensor fault detection},
\newblock in: \bibinfo{editor}{L.~Pancioni}, \bibinfo{editor}{F.~Schwenker},
  \bibinfo{editor}{E.~Trentin} (Eds.), \bibinfo{booktitle}{Artificial Neural
  Networks in Pattern Recognition}, \bibinfo{publisher}{Springer},
  \bibinfo{year}{2018}, pp. \bibinfo{pages}{320--332}.
\bibitem[{Chandola et~al.(2009)Chandola, Banerjee, and Kumar}]{Chandola}
\bibinfo{author}{V.~Chandola}, \bibinfo{author}{A.~Banerjee},
  \bibinfo{author}{V.~Kumar},
\newblock \bibinfo{title}{Anomaly detection: A survey},
\newblock \bibinfo{journal}{ACM Comput. Surv.} \bibinfo{volume}{41}
  (\bibinfo{year}{2009}) \bibinfo{pages}{15:1--15:58}.
  \DOIprefix\doi{10.1145/1541880.1541882}.
\bibitem[{Jurgovsky et~al.(2018)Jurgovsky, Granitzer, Ziegler, Calabretto,
  Portier, He, and Caelen}]{Jurgovsky}
\bibinfo{author}{J.~Jurgovsky}, \bibinfo{author}{M.~Granitzer},
  \bibinfo{author}{K.~Ziegler}, \bibinfo{author}{S.~Calabretto},
  \bibinfo{author}{P.-E. Portier}, \bibinfo{author}{L.~He},
  \bibinfo{author}{O.~Caelen},
\newblock \bibinfo{title}{Sequence classification for credit-card fraud
  detection},
\newblock \bibinfo{journal}{Expert Systems with Applications}
  \bibinfo{volume}{100} (\bibinfo{year}{2018}).
  \DOIprefix\doi{10.1016/j.eswa.2018.01.037}.
\bibitem[{Kirlidog and Asuk(2012)}]{kirlidog}
\bibinfo{author}{M.~Kirlidog}, \bibinfo{author}{C.~Asuk},
\newblock \bibinfo{title}{A fraud detection approach with data mining in health
  insurance},
\newblock \bibinfo{journal}{Procedia - Social and Behavioral Sciences}
  \bibinfo{volume}{62} (\bibinfo{year}{2012}) \bibinfo{pages}{989 -- 994}.
  \bibinfo{note}{World Conference on Business, Economics and Management
  (BEM-2012), May 4–6 2012, Antalya, Turkey}.
\bibitem[{Zhou et~al.(2018)Zhou, Cheng, Zhu, Guo, Zhou, Xu, Xue, and
  Zhang}]{zhou2018state}
\bibinfo{author}{X.~Zhou}, \bibinfo{author}{S.~Cheng},
  \bibinfo{author}{M.~Zhu}, \bibinfo{author}{C.~Guo},
  \bibinfo{author}{S.~Zhou}, \bibinfo{author}{P.~Xu}, \bibinfo{author}{Z.~Xue},
  \bibinfo{author}{W.~Zhang},
\newblock \bibinfo{title}{A state of the art survey of data mining-based fraud
  detection and credit scoring},
\newblock in: \bibinfo{booktitle}{MATEC Web of Conferences}, volume
  \bibinfo{volume}{189}, \bibinfo{organization}{EDP Sciences},
  \bibinfo{year}{2018}, p. \bibinfo{pages}{03002}.
\bibitem[{Phua et~al.(2010)Phua, Lee, Smith, and
  Gayler}]{phua2010comprehensive}
\bibinfo{author}{C.~Phua}, \bibinfo{author}{V.~Lee},
  \bibinfo{author}{K.~Smith}, \bibinfo{author}{R.~Gayler},
\newblock \bibinfo{title}{A comprehensive survey of data mining-based fraud
  detection research},
\newblock \bibinfo{journal}{arXiv preprint arXiv:1009.6119}
  (\bibinfo{year}{2010}).
\bibitem[{Wang and Xu(2018)}]{wang2018leveraging}
\bibinfo{author}{Y.~Wang}, \bibinfo{author}{W.~Xu},
\newblock \bibinfo{title}{Leveraging deep learning with lda-based text
  analytics to detect automobile insurance fraud},
\newblock \bibinfo{journal}{Decision Support Systems} \bibinfo{volume}{105}
  (\bibinfo{year}{2018}) \bibinfo{pages}{87--95}.
\bibitem[{Kim et~al.(2019)Kim, Kim, and Kim}]{kim2019fraud}
\bibinfo{author}{J.~Kim}, \bibinfo{author}{H.-J. Kim},
  \bibinfo{author}{H.~Kim},
\newblock \bibinfo{title}{Fraud detection for job placement using hierarchical
  clusters-based deep neural networks},
\newblock \bibinfo{journal}{Applied Intelligence}  (\bibinfo{year}{2019})
  \bibinfo{pages}{1--20}.
\bibitem[{Balasubramanian(2019)}]{balasubramanian2019ensemble}
\bibinfo{author}{M.~V. Balasubramanian}, \bibinfo{title}{Ensemble modeling \&
  prediction interpretability for insurance fraud claims classification}, Ph.D.
  thesis, Dublin Business School, \bibinfo{year}{2019}.
\bibitem[{Chen et~al.(2016)Chen, Tang, Sun, Chen, and Zhang}]{chen2016entity}
\bibinfo{author}{T.~Chen}, \bibinfo{author}{L.-A. Tang},
  \bibinfo{author}{Y.~Sun}, \bibinfo{author}{Z.~Chen},
  \bibinfo{author}{K.~Zhang}, \bibinfo{title}{Entity embedding-based anomaly
  detection for heterogeneous categorical events}, \bibinfo{year}{2016}.
  \href{http://arxiv.org/abs/1608.07502}{{\tt arXiv:1608.07502}}.
\bibitem[{Hu et~al.(2016)Hu, Aggarwal, Ma, and Huai}]{Hu2016AnEA}
\bibinfo{author}{R.~Hu}, \bibinfo{author}{C.~C. Aggarwal},
  \bibinfo{author}{S.~Ma}, \bibinfo{author}{J.~Huai},
\newblock \bibinfo{title}{An embedding approach to anomaly detection},
\newblock \bibinfo{journal}{2016 IEEE 32nd International Conference on Data
  Engineering (ICDE)}  (\bibinfo{year}{2016}) \bibinfo{pages}{385--396}.
\bibitem[{Rajaraman and Ullman(2011)}]{rajaraman2011mining}
\bibinfo{author}{A.~Rajaraman}, \bibinfo{author}{J.~D. Ullman},
  \bibinfo{title}{Mining of massive datasets}, \bibinfo{publisher}{Cambridge
  University Press}, \bibinfo{year}{2011}.
\bibitem[{Mikolov et~al.(2013)Mikolov, Sutskever, Chen, Corrado, and
  Dean}]{mikolov2013distributed}
\bibinfo{author}{T.~Mikolov}, \bibinfo{author}{I.~Sutskever},
  \bibinfo{author}{K.~Chen}, \bibinfo{author}{G.~S. Corrado},
  \bibinfo{author}{J.~Dean},
\newblock \bibinfo{title}{Distributed representations of words and phrases and
  their compositionality},
\newblock in: \bibinfo{booktitle}{Advances in neural information processing
  systems}, \bibinfo{year}{2013}, pp. \bibinfo{pages}{3111--3119}.
\bibitem[{Pennington et~al.(2014)Pennington, Socher, and
  Manning}]{pennington2014glove}
\bibinfo{author}{J.~Pennington}, \bibinfo{author}{R.~Socher},
  \bibinfo{author}{C.~Manning},
\newblock \bibinfo{title}{Glove: Global vectors for word representation},
\newblock in: \bibinfo{booktitle}{Proceedings of the 2014 conference on
  empirical methods in natural language processing (EMNLP)},
  \bibinfo{year}{2014}, pp. \bibinfo{pages}{1532--1543}.
\bibitem[{Arora et~al.(2016)Arora, Liang, and Ma}]{arora2016simple}
\bibinfo{author}{S.~Arora}, \bibinfo{author}{Y.~Liang},
  \bibinfo{author}{T.~Ma},
\newblock \bibinfo{title}{A simple but tough-to-beat baseline for sentence
  embeddings},
\newblock \bibinfo{journal}{ICLR}  (\bibinfo{year}{2016}).
\bibitem[{Wang et~al.(2016)Wang, Huang, Feng, Zhou, Gu, and Gao}]{wang2016cse}
\bibinfo{author}{Y.~Wang}, \bibinfo{author}{H.~Huang},
  \bibinfo{author}{C.~Feng}, \bibinfo{author}{Q.~Zhou},
  \bibinfo{author}{J.~Gu}, \bibinfo{author}{X.~Gao},
\newblock \bibinfo{title}{Cse: Conceptual sentence embeddings based on
  attention model},
\newblock in: \bibinfo{booktitle}{Proceedings of the 54th Annual Meeting of the
  Association for Computational Linguistics (Volume 1: Long Papers)},
  volume~\bibinfo{volume}{1}, \bibinfo{year}{2016}, pp.
  \bibinfo{pages}{505--515}.
\bibitem[{Kiros et~al.(2015)Kiros, Zhu, Salakhutdinov, Zemel, Urtasun,
  Torralba, and Fidler}]{kiros2015skip}
\bibinfo{author}{R.~Kiros}, \bibinfo{author}{Y.~Zhu}, \bibinfo{author}{R.~R.
  Salakhutdinov}, \bibinfo{author}{R.~Zemel}, \bibinfo{author}{R.~Urtasun},
  \bibinfo{author}{A.~Torralba}, \bibinfo{author}{S.~Fidler},
\newblock \bibinfo{title}{Skip-thought vectors},
\newblock in: \bibinfo{booktitle}{Advances in neural information processing
  systems}, \bibinfo{year}{2015}, pp. \bibinfo{pages}{3294--3302}.
\bibitem[{Krawczyk(2016)}]{Krawczyk2016}
\bibinfo{author}{B.~Krawczyk},
\newblock \bibinfo{title}{Learning from imbalanced data: open challenges and
  future directions},
\newblock \bibinfo{journal}{Progress in Artificial Intelligence}
  \bibinfo{volume}{5} (\bibinfo{year}{2016}) \bibinfo{pages}{221--232}.
  \URLprefix \url{https://doi.org/10.1007/s13748-016-0094-0}.
  \DOIprefix\doi{10.1007/s13748-016-0094-0}.
\bibitem[{Duman et~al.(2013)Duman, Buyukkaya, and Elikucuk}]{duman}
\bibinfo{author}{E.~Duman}, \bibinfo{author}{A.~Buyukkaya},
  \bibinfo{author}{I.~Elikucuk},
\newblock \bibinfo{title}{A novel and successful credit card fraud detection
  system implemented in a turkish bank},
\newblock in: \bibinfo{booktitle}{2013 IEEE 13th International Conference on
  Data Mining Workshops}, \bibinfo{organization}{IEEE}, \bibinfo{year}{2013},
  pp. \bibinfo{pages}{162--171}.
\bibitem[{Erofeev et~al.(2015)Erofeev, Burnaev, and Papanov}]{Imbalance2015}
\bibinfo{author}{P.~Erofeev}, \bibinfo{author}{E.~Burnaev},
  \bibinfo{author}{A.~Papanov},
\newblock \bibinfo{title}{Influence of resampling on accuracy of imbalanced
  classification},
\newblock in: \bibinfo{booktitle}{Proc. SPIE}, volume \bibinfo{volume}{9875},
  \bibinfo{year}{2015}, pp. \bibinfo{pages}{9875--9875--5}.
  \DOIprefix\doi{10.1117/12.2228523}.
\bibitem[{Sahin et~al.(2013)Sahin, Bulkan, and Duman}]{Sahin2013}
\bibinfo{author}{Y.~Sahin}, \bibinfo{author}{S.~Bulkan},
  \bibinfo{author}{E.~Duman},
\newblock \bibinfo{title}{A cost-sensitive decision tree approach for fraud
  detection},
\newblock \bibinfo{journal}{Expert Syst. Appl.} \bibinfo{volume}{40}
  (\bibinfo{year}{2013}) \bibinfo{pages}{5916--5923}.
\bibitem[{Seeja and Zareapoor(2014)}]{fraudminer}
\bibinfo{author}{K.~Seeja}, \bibinfo{author}{M.~Zareapoor},
\newblock \bibinfo{title}{Fraudminer: A novel credit card fraud detection model
  based on frequent itemset mining},
\newblock \bibinfo{journal}{TheScientificWorldJournal} \bibinfo{volume}{2014}
  (\bibinfo{year}{2014}) \bibinfo{pages}{252797}.
  \DOIprefix\doi{10.1155/2014/252797}.
\bibitem[{Chawla et~al.(2002)Chawla, Bowyer, Hall, and
  Kegelmeyer}]{chawla2002smote}
\bibinfo{author}{N.~V. Chawla}, \bibinfo{author}{K.~W. Bowyer},
  \bibinfo{author}{L.~O. Hall}, \bibinfo{author}{W.~P. Kegelmeyer},
\newblock \bibinfo{title}{Smote: synthetic minority over-sampling technique},
\newblock \bibinfo{journal}{Journal of artificial intelligence research}
  \bibinfo{volume}{16} (\bibinfo{year}{2002}) \bibinfo{pages}{321--357}.
\bibitem[{S{\'a}ez et~al.(2015)S{\'a}ez, Luengo, Stefanowski, and
  Herrera}]{saez2015smote}
\bibinfo{author}{J.~A. S{\'a}ez}, \bibinfo{author}{J.~Luengo},
  \bibinfo{author}{J.~Stefanowski}, \bibinfo{author}{F.~Herrera},
\newblock \bibinfo{title}{Smote--ipf: Addressing the noisy and borderline
  examples problem in imbalanced classification by a re-sampling method with
  filtering},
\newblock \bibinfo{journal}{Information Sciences} \bibinfo{volume}{291}
  (\bibinfo{year}{2015}) \bibinfo{pages}{184--203}.
\bibitem[{Smolyakov et~al.(2019)Smolyakov, Korotin, Erofeev, Papanov, and
  Burnaev}]{Imbalance2019}
\bibinfo{author}{D.~Smolyakov}, \bibinfo{author}{A.~Korotin},
  \bibinfo{author}{P.~Erofeev}, \bibinfo{author}{A.~Papanov},
  \bibinfo{author}{E.~Burnaev},
\newblock \bibinfo{title}{Meta-learning for resampling recommendation systems},
\newblock in: \bibinfo{booktitle}{Proc. SPIE 11041, Eleventh International
  Conference on Machine Vision (ICMV 2018), 110411S (15 March 2019)},
  \bibinfo{year}{2019}.
\bibitem[{Makridakis(2017)}]{makridakis2017forthcoming}
\bibinfo{author}{S.~Makridakis},
\newblock \bibinfo{title}{The forthcoming artificial intelligence (ai)
  revolution: Its impact on society and firms},
\newblock \bibinfo{journal}{Futures} \bibinfo{volume}{90}
  (\bibinfo{year}{2017}) \bibinfo{pages}{46--60}.
\bibitem[{LeCun et~al.(2015)LeCun, Bengio, and Hinton}]{lecun2015deep}
\bibinfo{author}{Y.~LeCun}, \bibinfo{author}{Y.~Bengio},
  \bibinfo{author}{G.~Hinton},
\newblock \bibinfo{title}{Deep learning},
\newblock \bibinfo{journal}{nature} \bibinfo{volume}{521}
  (\bibinfo{year}{2015}) \bibinfo{pages}{436}.
\bibitem[{Vasilache et~al.(2014)Vasilache, Johnson, Mathieu, Chintala,
  Piantino, and LeCun}]{vasilache2014fast}
\bibinfo{author}{N.~Vasilache}, \bibinfo{author}{J.~Johnson},
  \bibinfo{author}{M.~Mathieu}, \bibinfo{author}{S.~Chintala},
  \bibinfo{author}{S.~Piantino}, \bibinfo{author}{Y.~LeCun},
\newblock \bibinfo{title}{Fast convolutional nets with fbfft: A gpu performance
  evaluation},
\newblock \bibinfo{journal}{arXiv preprint arXiv:1412.7580}
  (\bibinfo{year}{2014}).
\bibitem[{Russakovsky et~al.(2015)Russakovsky, Deng, Su, Krause, Satheesh, Ma,
  Huang, Karpathy, Khosla, Bernstein, Berg, and Fei-Fei}]{ILSVRC15}
\bibinfo{author}{O.~Russakovsky}, \bibinfo{author}{J.~Deng},
  \bibinfo{author}{H.~Su}, \bibinfo{author}{J.~Krause},
  \bibinfo{author}{S.~Satheesh}, \bibinfo{author}{S.~Ma},
  \bibinfo{author}{Z.~Huang}, \bibinfo{author}{A.~Karpathy},
  \bibinfo{author}{A.~Khosla}, \bibinfo{author}{M.~Bernstein},
  \bibinfo{author}{A.~C. Berg}, \bibinfo{author}{L.~Fei-Fei},
\newblock \bibinfo{title}{{ImageNet Large Scale Visual Recognition Challenge}},
\newblock \bibinfo{journal}{International Journal of Computer Vision (IJCV)}
  \bibinfo{volume}{115} (\bibinfo{year}{2015}) \bibinfo{pages}{211--252}.
  \DOIprefix\doi{10.1007/s11263-015-0816-y}.
\bibitem[{Amodei et~al.(2016)Amodei, Ananthanarayanan, Anubhai, Bai,
  Battenberg, Case, Casper, Catanzaro, Cheng, Chen et~al.}]{amodei2016deep}
\bibinfo{author}{D.~Amodei}, \bibinfo{author}{S.~Ananthanarayanan},
  \bibinfo{author}{R.~Anubhai}, \bibinfo{author}{J.~Bai},
  \bibinfo{author}{E.~Battenberg}, \bibinfo{author}{C.~Case},
  \bibinfo{author}{J.~Casper}, \bibinfo{author}{B.~Catanzaro},
  \bibinfo{author}{Q.~Cheng}, \bibinfo{author}{G.~Chen}, et~al.,
\newblock \bibinfo{title}{Deep speech 2: End-to-end speech recognition in
  english and mandarin},
\newblock in: \bibinfo{booktitle}{International conference on machine
  learning}, \bibinfo{year}{2016}, pp. \bibinfo{pages}{173--182}.
\bibitem[{Young et~al.(2018)Young, Hazarika, Poria, and
  Cambria}]{young2018recent}
\bibinfo{author}{T.~Young}, \bibinfo{author}{D.~Hazarika},
  \bibinfo{author}{S.~Poria}, \bibinfo{author}{E.~Cambria},
\newblock \bibinfo{title}{Recent trends in deep learning based natural language
  processing},
\newblock \bibinfo{journal}{ieee Computational intelligenCe magazine}
  \bibinfo{volume}{13} (\bibinfo{year}{2018}) \bibinfo{pages}{55--75}.
\bibitem[{Hamilton et~al.(2017)Hamilton, Ying, and
  Leskovec}]{hamilton2017representation}
\bibinfo{author}{W.~L. Hamilton}, \bibinfo{author}{R.~Ying},
  \bibinfo{author}{J.~Leskovec},
\newblock \bibinfo{title}{Representation learning on graphs: Methods and
  applications},
\newblock \bibinfo{journal}{arXiv preprint arXiv:1709.05584}
  (\bibinfo{year}{2017}).
\bibitem[{Mikolov et~al.(2013)Mikolov, Chen, Corrado, and
  Dean}]{mikolov2013efficient}
\bibinfo{author}{T.~Mikolov}, \bibinfo{author}{K.~Chen},
  \bibinfo{author}{G.~Corrado}, \bibinfo{author}{J.~Dean},
\newblock \bibinfo{title}{Efficient estimation of word representations in
  vector space},
\newblock \bibinfo{journal}{arXiv preprint arXiv:1301.3781}
  (\bibinfo{year}{2013}).
\bibitem[{Bartunov et~al.(2016)Bartunov, Kondrashkin, Osokin, and
  Vetrov}]{bartunov2016breaking}
\bibinfo{author}{S.~Bartunov}, \bibinfo{author}{D.~Kondrashkin},
  \bibinfo{author}{A.~Osokin}, \bibinfo{author}{D.~Vetrov},
\newblock \bibinfo{title}{Breaking sticks and ambiguities with adaptive
  skip-gram},
\newblock in: \bibinfo{booktitle}{Artificial Intelligence and Statistics},
  \bibinfo{year}{2016}, pp. \bibinfo{pages}{130--138}.
\bibitem[{Collobert et~al.(2011)Collobert, Weston, Bottou, Karlen, Kavukcuoglu,
  and Kuksa}]{collobert2011natural}
\bibinfo{author}{R.~Collobert}, \bibinfo{author}{J.~Weston},
  \bibinfo{author}{L.~Bottou}, \bibinfo{author}{M.~Karlen},
  \bibinfo{author}{K.~Kavukcuoglu}, \bibinfo{author}{P.~Kuksa},
\newblock \bibinfo{title}{Natural language processing (almost) from scratch},
\newblock \bibinfo{journal}{Journal of Machine Learning Research}
  \bibinfo{volume}{12} (\bibinfo{year}{2011}) \bibinfo{pages}{2493--2537}.
\bibitem[{Zou et~al.(2013)Zou, Socher, Cer, and Manning}]{zou2013bilingual}
\bibinfo{author}{W.~Y. Zou}, \bibinfo{author}{R.~Socher},
  \bibinfo{author}{D.~Cer}, \bibinfo{author}{C.~D. Manning},
\newblock \bibinfo{title}{Bilingual word embeddings for phrase-based machine
  translation},
\newblock in: \bibinfo{booktitle}{Proceedings of the 2013 Conference on
  Empirical Methods in Natural Language Processing}, \bibinfo{year}{2013}, pp.
  \bibinfo{pages}{1393--1398}.
\bibitem[{Ma and Hovy(2016)}]{ma2016end}
\bibinfo{author}{X.~Ma}, \bibinfo{author}{E.~Hovy},
\newblock \bibinfo{title}{End-to-end sequence labeling via bi-directional
  lstm-cnns-crf},
\newblock in: \bibinfo{booktitle}{Proceedings of the 54th Annual Meeting of the
  Association for Computational Linguistics}, \bibinfo{year}{2016}, p.
  \bibinfo{pages}{1064–1074}.
\bibitem[{Vishwanathan et~al.(2010)Vishwanathan, Schraudolph, Kondor, and
  Borgwardt}]{rwkernel:10}
\bibinfo{author}{S.~V.~N. Vishwanathan}, \bibinfo{author}{N.~N. Schraudolph},
  \bibinfo{author}{R.~Kondor}, \bibinfo{author}{K.~M. Borgwardt},
\newblock \bibinfo{title}{Graph kernels},
\newblock \bibinfo{journal}{J. Mach. Learn. Res.} \bibinfo{volume}{11}
  (\bibinfo{year}{2010}) \bibinfo{pages}{1201--1242}.
\bibitem[{Haussler(1999)}]{rconvolution}
\bibinfo{author}{D.~Haussler}, \bibinfo{title}{Convolution Kernels on Discrete
  Structures}, \bibinfo{type}{Technical Report}, University of California at
  Santa Cruz, \bibinfo{year}{1999}.
\bibitem[{Yanardag and Vishwanathan(2015)}]{deepgraph}
\bibinfo{author}{P.~Yanardag}, \bibinfo{author}{S.~V.~N. Vishwanathan},
\newblock \bibinfo{title}{Deep graph kernels},
\newblock in: \bibinfo{booktitle}{Proceedings of the 21th {ACM} {SIGKDD}
  International Conference on Knowledge Discovery and Data Mining, Sydney, NSW,
  Australia, August 10-13, 2015}, \bibinfo{year}{2015}, pp.
  \bibinfo{pages}{1365--1374}.
\bibitem[{Niepert et~al.(2016)Niepert, Ahmed, and Kutzkov}]{learncnn:16}
\bibinfo{author}{M.~Niepert}, \bibinfo{author}{M.~Ahmed},
  \bibinfo{author}{K.~Kutzkov},
\newblock \bibinfo{title}{Learning convolutional neural networks for graphs},
\newblock in: \bibinfo{booktitle}{Proceedings of the 33nd International
  Conference on Machine Learning, {ICML} 2016, New York City, NY, USA, June
  19-24, 2016}, \bibinfo{year}{2016}, pp. \bibinfo{pages}{2014--2023}.
\bibitem[{Ivanov and Burnaev(2018)}]{AWE}
\bibinfo{author}{S.~Ivanov}, \bibinfo{author}{E.~Burnaev},
\newblock \bibinfo{title}{Anonymous walk embeddings},
\newblock in: \bibinfo{editor}{J.~Dy}, \bibinfo{editor}{A.~Krause} (Eds.),
  \bibinfo{booktitle}{Proceedings of the 35th International Conference on
  Machine Learning}, volume~\bibinfo{volume}{80} of
  \textit{\bibinfo{series}{Proceedings of Machine Learning Research}},
  \bibinfo{publisher}{PMLR}, \bibinfo{year}{2018}, pp.
  \bibinfo{pages}{2186--2195}.
\bibitem[{Ivanov et~al.(2018)Ivanov, Durasov, and Burnaev}]{InfluenceSet2018}
\bibinfo{author}{S.~Ivanov}, \bibinfo{author}{N.~Durasov},
  \bibinfo{author}{E.~Burnaev},
\newblock \bibinfo{title}{Learning node embeddings for influence set
  completion},
\newblock in: \bibinfo{booktitle}{Proc. of IEEE International Conference on
  Data Mining Workshops (ICDMW)}, \bibinfo{year}{2018}, pp.
  \bibinfo{pages}{1034--1037}.
\bibitem[{{\v R}eh{\r u}{\v r}ek and Sojka(2010)}]{rehurek_lrec}
\bibinfo{author}{R.~{\v R}eh{\r u}{\v r}ek}, \bibinfo{author}{P.~Sojka},
\newblock \bibinfo{title}{{Software Framework for Topic Modelling with Large
  Corpora}},
\newblock in: \bibinfo{booktitle}{{Proceedings of the LREC 2010 Workshop on New
  Challenges for NLP Frameworks}}, \bibinfo{publisher}{ELRA},
  \bibinfo{address}{Valletta, Malta}, \bibinfo{year}{2010}, pp.
  \bibinfo{pages}{45--50}.
  \bibinfo{note}{\url{http://is.muni.cz/publication/884893/en}}.
\bibitem[{Fern{\'a}ndez-Delgado et~al.(2014)Fern{\'a}ndez-Delgado, Cernadas,
  Barro, and Amorim}]{fernandez2014we}
\bibinfo{author}{M.~Fern{\'a}ndez-Delgado}, \bibinfo{author}{E.~Cernadas},
  \bibinfo{author}{S.~Barro}, \bibinfo{author}{D.~Amorim},
\newblock \bibinfo{title}{Do we need hundreds of classifiers to solve real
  world classification problems},
\newblock \bibinfo{journal}{J. Mach. Learn. Res} \bibinfo{volume}{15}
  (\bibinfo{year}{2014}) \bibinfo{pages}{3133--3181}.
\bibitem[{Chen and Guestrin(2016)}]{chen2016xgboost}
\bibinfo{author}{T.~Chen}, \bibinfo{author}{C.~Guestrin},
\newblock \bibinfo{title}{{XGBoost}: A scalable tree boosting system},
\newblock in: \bibinfo{booktitle}{Proc. of the 22nd ACM SIGKDD int. conf. on
  knowledge discovery and data mining}, \bibinfo{organization}{ACM},
  \bibinfo{year}{2016}, pp. \bibinfo{pages}{785--794}.
\bibitem[{Kozlovskaia and Zaytsev(2017)}]{kozlovskaya2017deepboost}
\bibinfo{author}{N.~Kozlovskaia}, \bibinfo{author}{A.~Zaytsev},
\newblock \bibinfo{title}{Deep ensembles for imbalanced classification},
\newblock in: \bibinfo{booktitle}{Machine Learning and Applications (ICMLA),
  2017 16th IEEE International Conference on}, \bibinfo{year}{2017}, pp.
  \bibinfo{pages}{908--913}.
\bibitem[{Shen et~al.(2018)Shen, Wang, Wang, Min, Su, Zhang, Li, Henao, and
  Carin}]{swem}
\bibinfo{author}{D.~Shen}, \bibinfo{author}{G.~Wang},
  \bibinfo{author}{W.~Wang}, \bibinfo{author}{M.~R. Min},
  \bibinfo{author}{Q.~Su}, \bibinfo{author}{Y.~Zhang}, \bibinfo{author}{C.~Li},
  \bibinfo{author}{R.~Henao}, \bibinfo{author}{L.~Carin},
\newblock \bibinfo{title}{Baseline needs more love: On simple
  word-embedding-based models and associated pooling mechanisms},
\newblock \bibinfo{journal}{CoRR} \bibinfo{volume}{abs/1805.09843}
  (\bibinfo{year}{2018}). \URLprefix \url{http://arxiv.org/abs/1805.09843}.
  \href{http://arxiv.org/abs/1805.09843}{{\tt arXiv:1805.09843}}.
\bibitem[{Farbmacher et~al.(2019)Farbmacher, Loew, and Spindler}]{FLS}
\bibinfo{author}{H.~Farbmacher}, \bibinfo{author}{L.~Loew},
  \bibinfo{author}{M.~Spindler}, \bibinfo{title}{An Explainable Attention
  Network for Fraud Detection in Claims Management}, \bibinfo{type}{Technical
  Report}, University og Hamburg, \bibinfo{year}{2019}.
\bibitem[{Montemurro(2001)}]{montemurro2001beyond}
\bibinfo{author}{M.~A. Montemurro},
\newblock \bibinfo{title}{Beyond the zipf--mandelbrot law in quantitative
  linguistics},
\newblock \bibinfo{journal}{Physica A: Statistical Mechanics and its
  Applications} \bibinfo{volume}{300} (\bibinfo{year}{2001})
  \bibinfo{pages}{567--578}.
\bibitem[{Yin and Shen(2018)}]{yin2018dimensionality}
\bibinfo{author}{Z.~Yin}, \bibinfo{author}{Y.~Shen},
\newblock \bibinfo{title}{On the dimensionality of word embedding},
\newblock in: \bibinfo{booktitle}{Advances in Neural Information Processing
  Systems}, \bibinfo{year}{2018}, pp. \bibinfo{pages}{887--898}.
\bibitem[{Lema{{\^i}}tre et~al.(2017)Lema{{\^i}}tre, Nogueira, and
  Aridas}]{pacpacpac}
\bibinfo{author}{G.~Lema{{\^i}}tre}, \bibinfo{author}{F.~Nogueira},
  \bibinfo{author}{C.~K. Aridas},
\newblock \bibinfo{title}{Imbalanced-learn: A python toolbox to tackle the
  curse of imbalanced datasets in machine learning},
\newblock \bibinfo{journal}{Journal of Machine Learning Research}
  \bibinfo{volume}{18} (\bibinfo{year}{2017}) \bibinfo{pages}{1--5}. \URLprefix
  \url{http://jmlr.org/papers/v18/16-365}.
\bibitem[{Papernot et~al.(2017)Papernot, McDaniel, Goodfellow, Jha, Celik, and
  Swami}]{papernot2017practical}
\bibinfo{author}{N.~Papernot}, \bibinfo{author}{P.~McDaniel},
  \bibinfo{author}{I.~Goodfellow}, \bibinfo{author}{S.~Jha},
  \bibinfo{author}{Z.~B. Celik}, \bibinfo{author}{A.~Swami},
\newblock \bibinfo{title}{Practical black-box attacks against machine
  learning},
\newblock in: \bibinfo{booktitle}{Proceedings of the 2017 ACM on Asia
  conference on computer and communications security},
  \bibinfo{organization}{ACM}, \bibinfo{year}{2017}, pp.
  \bibinfo{pages}{506--519}.
\bibitem[{Saltelli et~al.(2004)Saltelli, Tarantola, Campolongo, and
  Ratto}]{saltelli2004sensitivity}
\bibinfo{author}{A.~Saltelli}, \bibinfo{author}{S.~Tarantola},
  \bibinfo{author}{F.~Campolongo}, \bibinfo{author}{M.~Ratto},
\newblock \bibinfo{title}{Sensitivity analysis in practice: a guide to
  assessing scientific models},
\newblock \bibinfo{journal}{Chichester, England}  (\bibinfo{year}{2004}).

\end{thebibliography}

\end{document}